\documentclass{article}

\usepackage{PRIMEarxiv}

\usepackage[utf8]{inputenc} 
\usepackage[T1]{fontenc}    
\usepackage{hyperref}       
\usepackage{url}            
\usepackage{booktabs}       
\usepackage{amsfonts}       
\usepackage{nicefrac}       
\usepackage{microtype}      
\usepackage{lipsum}
\usepackage{fancyhdr}       
\usepackage{graphicx}       
\usepackage{multirow}

\usepackage{booktabs}
\usepackage{tabularx}
\usepackage{xcolor}
\usepackage{marvosym}
\usepackage{ifsym}
\usepackage{subcaption}

\usepackage{amsmath}
\usepackage{bookmark}
\graphicspath{{media/}}     

\title{Enhancing Visual Grounding and Generalization: A Multi-Task Cycle Training Approach for Vision-Language Models}

\author{Xiaoyu~Yang \textsuperscript{1,2}, 
Lijian~Xu \textsuperscript{2,3 ({\Letter}) },
Hao~Sun \textsuperscript{3},
Hongsheng~Li \textsuperscript{3,4}\&
Shaoting~Zhang \textsuperscript{2}\\
\textsuperscript{1} College of Electronics and Information Engineering, Tongji University, Shanghai \\
\textsuperscript{2} Shanghai Artificial Intelligence Laboratory, Shanghai \\
\textsuperscript{3} Centre for Perceptual and Interactive Intelligence, the Chinese University of Hong Kong, Hong Kong \\
\textsuperscript{4} Department of Electronic Engineering, the Chinese University of Hong Kong, Hong Kong \\
}

\begin{document}
\maketitle

\begin{abstract}
  Visual grounding occupies a pivotal position in multi-modality vision-language models. 
  However, current vision-language models concentrate on comprehending images, ignoring the human-computer interaction with multi-tasks instructions,  thereby imposing limitations on their versatility and depth of responses.
  In this study, we propose ViLaM, a large multi-modality model, that supports multi-tasks of visual grounding using the cycle training strategy, with abundant interaction instructions.
  The cycle training between referring expression generation (REG) and referring expression comprehension (REC) is introduced. It enhances the consistency between visual location and referring expressions, and addresses the need for high-quality, multi-tasks visual grounding datasets. 
  Moreover, multi-tasks of visual grounding are promoted in our model, contributed by the cycle training strategy. The multi-tasks in REC encompass a range of granularities, from region-level to pixel-level, which include referring bbox detection, referring keypoints detection, and referring image segmentation. In REG, referring region classification determines the fine-grained category of the target, while referring region captioning generates a comprehensive description. Meanwhile, all tasks participate in the joint training, synergistically enhancing one another and collectively improving the overall performance of the model.
  Furthermore, leveraging the capabilities of large language models, ViLaM extends a wide range of instructions, thereby significantly enhancing its generalization and interaction potentials. It is particularly advantageous in domains beyond natural images, such as the medical field.
  Extensive public datasets corroborate the superior capabilities of our model in visual grounding with muti-tasks. Additionally, validating its robust generalization, ViLaM is validated under open-set and few-shot scenarios. Especially in the medical field, our model demonstrates cross-domain robust generalization capabilities.
  Furthermore, we contribute a visual grounding dataset, especially with multi-tasks. To support and encourage the community focused on visual grounding, we have made both the dataset and our code public: \url{https://github.com/AnonymGiant/ViLaM}.
\end{abstract}


\section{Introduction}



Visual grounding serves as a vital bridge that facilitates communication between AI models and the world, representing a significant milestone in the quest for achieving general intelligence. However, traditional approaches to visual grounding predominantly prioritize image comprehension, striving to extract image features with greater accuracy and align them with textual information. Ignoring the wealth of referential information present in the texts of referring expressions leads to the lack of diverse instructions and limited versatility and generalization.

In contrast to traditional methods, the advent of Large Language Models (LLMs) has significantly mitigated this issue, and spawned rethinking about the vision-language models. Leveraging extensive text data, LLMs have attained remarkable proficiency in generating human-like responses and addressing a wide array of tasks. This breakthrough ushers in a new paradigm for human-computer interaction. Building upon LLMs, large multi-modality models like BLIP2 \cite{li2023blip2} and MiniGPT4 \cite{zhu2023minigpt4} utilize a pre-trained image encoder and a text encoder, then align vision-language features with simple linear layers. 
These models demonstrate impressive joint understanding capabilities of language and images, allowing users to give instructions in natural language to perform specific tasks.
Nonetheless, the visual grounding task of aligning visual localization with referring expressions remains a challenge for large multi-modality models, particularly when dealing with localization spanning various granularities, such as bounding boxes, keypoints, and segmentated polygons that range from region-level to pixel-level precision.  In particular, pre-trained LLMs mainly developed for specific language tasks often struggle with image-processing tasks, arduous to ensure the consistency between visual localization and referring expression.

\begin{figure*}[htbp]
    \centering
    \includegraphics[width=\textwidth]{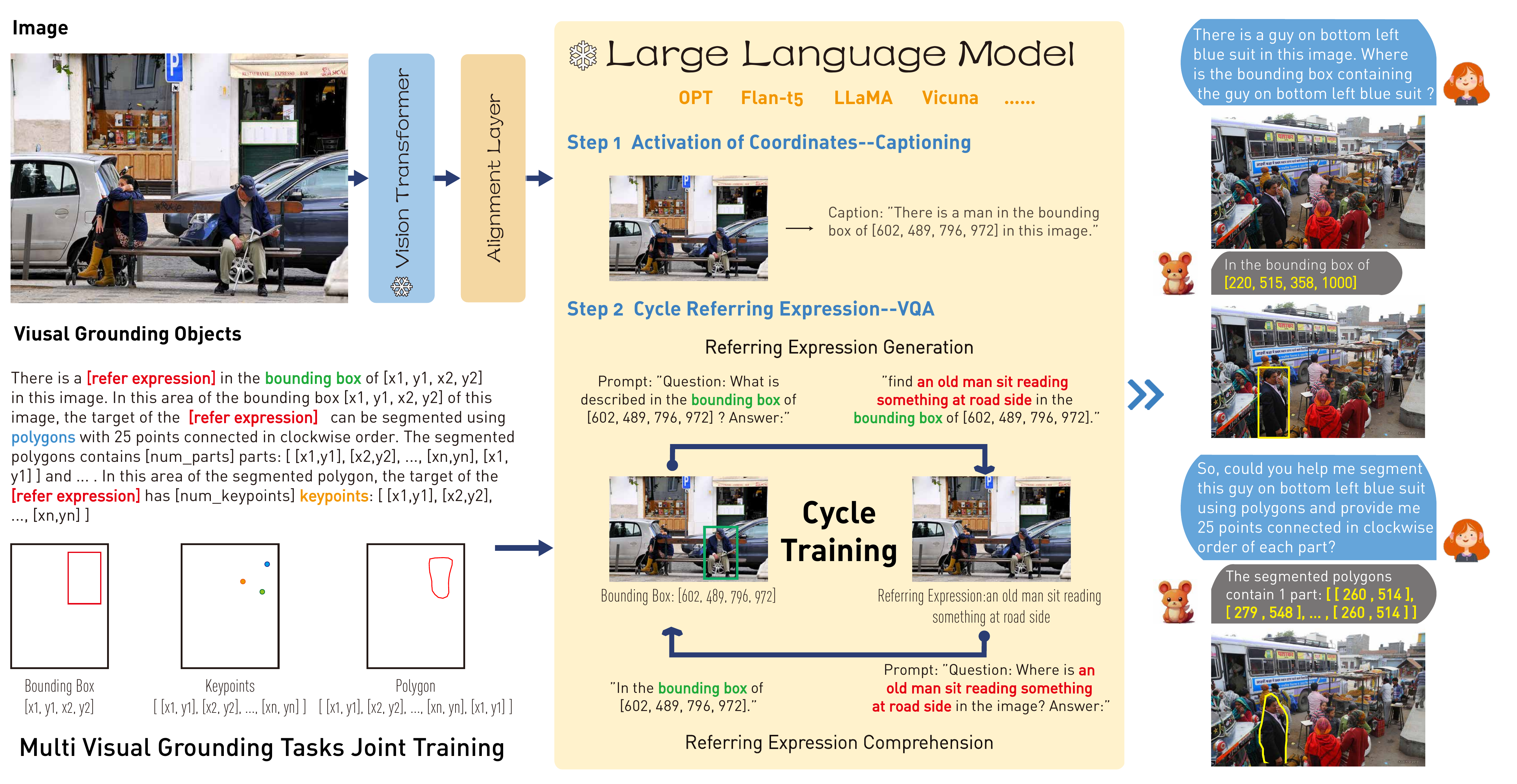}
    \caption{The workflow of our methodology. 
    We design the strategy of cycle training for referring expressions to activate the orientation capability of the vision-language model. The coordinates of the location will cycle through two subtasks: referring expression generation and referring expression comprehension to enhance the consistency between visual location and referring expressions.
    Furthermore, the joint training of multi-tasks of visual grounding is presented to improve the visual grounding at various levels of granularity, including referring bounding box detection, referring keypoints detection and referring image segmentation. 
    The pre-trained visual model and large language model are frozen in the training.}
    \label{fig:model}
\end{figure*}

In this study, we introduce ViLaM, a large multi-modality model, that supports multi-tasks of visual grounding using cycle training strategy, with abundant instructions. First, to enhance the visual grounding capabilities of large multi-modality models, the cycle training strategy is introduced, wherein referring expression generation and referring expression comprehension are cycled to enhance the consistency between visual location and referring expressions. Furthermore, the cycle training approach also contributes to performance improvement through the additional datasets without referring expressions or visual locations. The pairs of referring expressions and visual locations can be generated using the cycle training, thereby harnessing the power of big data to drive the large model. 
Second, multi-tasks of visual grounding are supported and participate in the joint training in our model. The visual localization capability of the ViLam covers various granularities from the region-level to pixel-level, including bounding box (bbox), keypoints and segmented polygons. Regarding captioning tasks, our proposed model excels in producing fine-grained categorization of the identified targets and generating comprehensive descriptions that encapsulate the attributes of the pointed target. Thereby, the joint training of multiple tasks enhances the consistency between visual localization and referring expressions from various perspectives, leading to improved performance across different tasks and the overall model. 
Meanwhile, leveraging the power of LLMs, our approach supports a wide array of abundant instruction prompts, significantly enhancing the human-computer interaction and generalization capabilities of our model. Especially in cross-domain tasks, such as the medical field, our model possesses the capability to effectively accomplish novel tasks based on given instructions.
Finally, recognizing the demand for higher-quality visual grounding data that encompasses diverse forms of visual location, we developed a dataset called VGCoco. It contains about 240K images with grounding annotations varying from region-level to pixel-level, including bboxes, keypoints and segmented polygons. In order to foster and support the visual grounding community, we have made our dataset and code publicly available.

To summarize, our contributions are four-fold: 
(1) We design the cycle training, enhancing the consistency between visual localization and referring expressions, and satisfying the requirements of paired visual grounding datasets for the large model training, both in quantity and quality.
(2) Incorporating the LLMs with abundant instructions, multi-tasks of visual grounding are supported in the ViLaM and participate in the joint training, enhancing the capabilities of visual grounding and generalization.
(3) We assess the performance of ViLaM in extensive public datasets with multi-tasks, demonstrating its superior capabilities of visual grounding. Besides, the generalization of ViLaM is verified under the open-set or few-shot seniors, especially in cross-domain, such as the medical field. 
(4)  To foster and empower the community, we contribute a visual grounding dataset with multi-tasks annotations. The dataset and our code are public.

\section{Related Work}

\subsection{LLMs and Multi-modality Pre-training}


Large Language Models (LLMs) have recently significantly impacted the field of natural language processing. Through alignment techniques such as supervised learning and reinforcement learning with human feedback, LLMs can effectively generalize to perform a wide range of tasks, even with limited training data. 
A remarkable application of LLM is ChatGPT, which presents an amazing ability to interact with humans. OpenAI's ChatGPT and GPT4 are prime examples of the impact that AI can have, and there have been extensive open-source efforts to replicate their success, such as OPT \cite{zhang2022opt}, BLOOM \cite{scao2022bloom}, PALM \cite{chowdhery2022palm}, LLaMA \cite{touvron2023llama}.

Multi-modality models have further promoted the development of the vision-language model \cite{radford2021learning,li2022grounded,alayrac2022flamingo,li2023blip2,zhu2023minigpt4,liu2023improved,chen2023pali}. 
GPT-4V \cite{GPT-4V,liu2023improved}
has recently shown unprecedented ability in understanding and processing an
arbitrary mix of input images and texts. On the other hand, preliminary experiments show that visual grounding accuracy is still limited in the comprehensive scene, like the medical field.

\subsection{Visual Grounding}

\noindent\textbf{Referring Expression Comprehension}
Early pioneers typically used a two-stage approach to tackle visual grounding tasks. The initial step involves extracting interest regions, which are subsequently prioritized based on their similarity scores with the language query  \cite{gan2020large,yangDynamicGraphAttention2019,liuImprovingReferringExpression2019,liuLearningAssembleNeural2019}. 
Another line of work advocates a one-stage pipeline based on dense anchors \cite{chen2020uniter,yang2020improving,kamath2021mdetr,yang2022unitab,Deng_2021_ICCV}.  
Other Transformer models like SeqTR \cite{Zhu_2022},  VGTR \cite{du2022visual} and PolyFormer \cite{liuPolyFormerReferringImage2023a} in Vision-Language Tasks are subsequently proposed for the visual grounding task and achieved satisfactory performance.





\noindent\textbf{Generalist Model}
Recently, the potential of generalist models has been increasingly explored, garnering considerable attention from the research community. Among these, OFA \cite{wang2022ofa} integrates a diverse set of cross-modal and uni-modal tasks within a simple sequence-to-sequence learning framework. It adheres to instruction-based learning in both pre-training and fine-tuning stages, negating the need for additional task-specific layers for downstream tasks.
Besides, mPLUG-2 \cite{xuMPLUG2ModularizedMultimodal2023} presents a multi-module composition network that utilizes shared universal modules for modality collaboration and separates distinct modality modules to address modality entanglement.

\begin{figure*}[htbp]
    \centering
    \begin{subfigure}[t]{0.33\textwidth}
        \centering
        \includegraphics[width=0.99\textwidth]{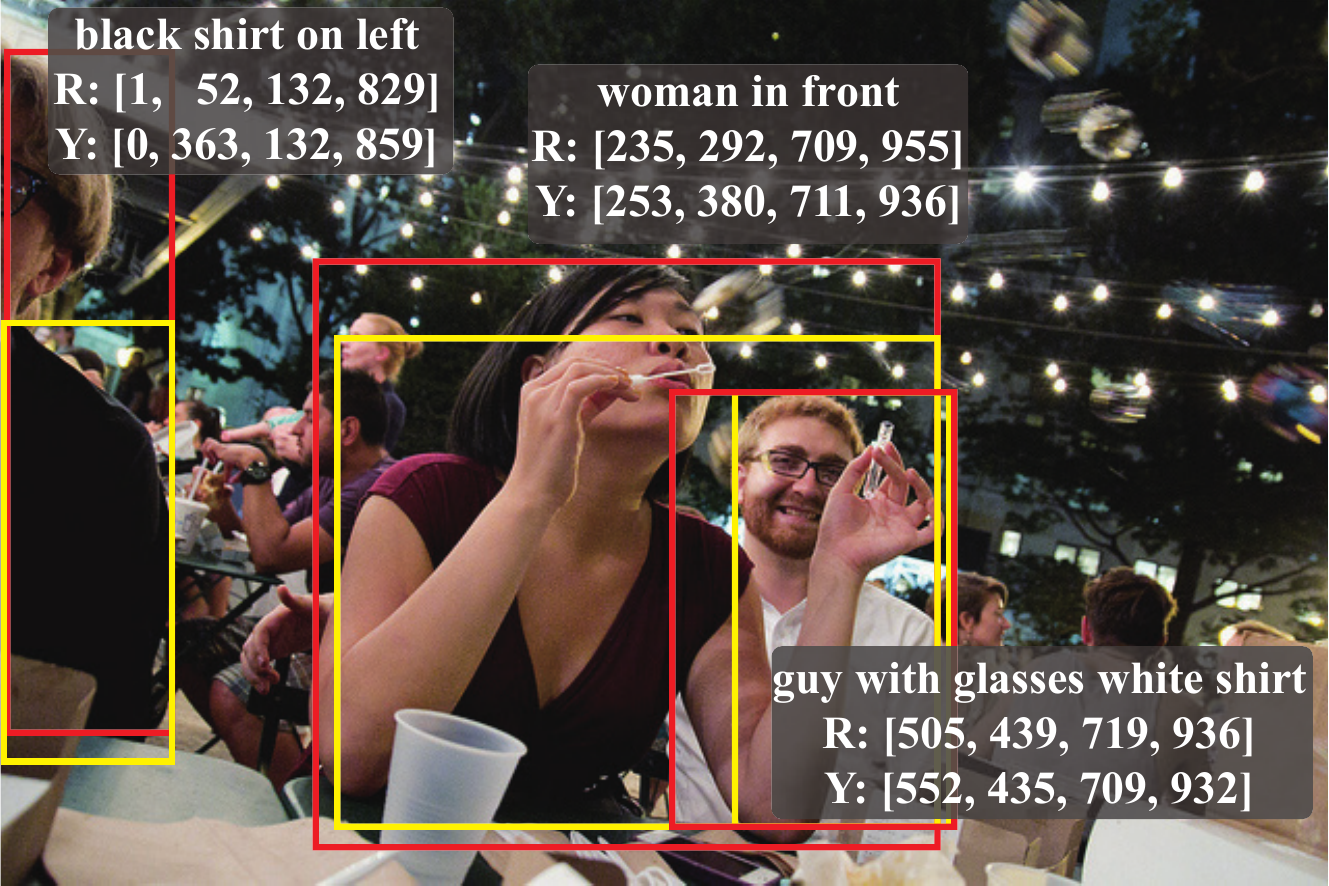}
        Persons\\

        \includegraphics[width=0.99\textwidth]{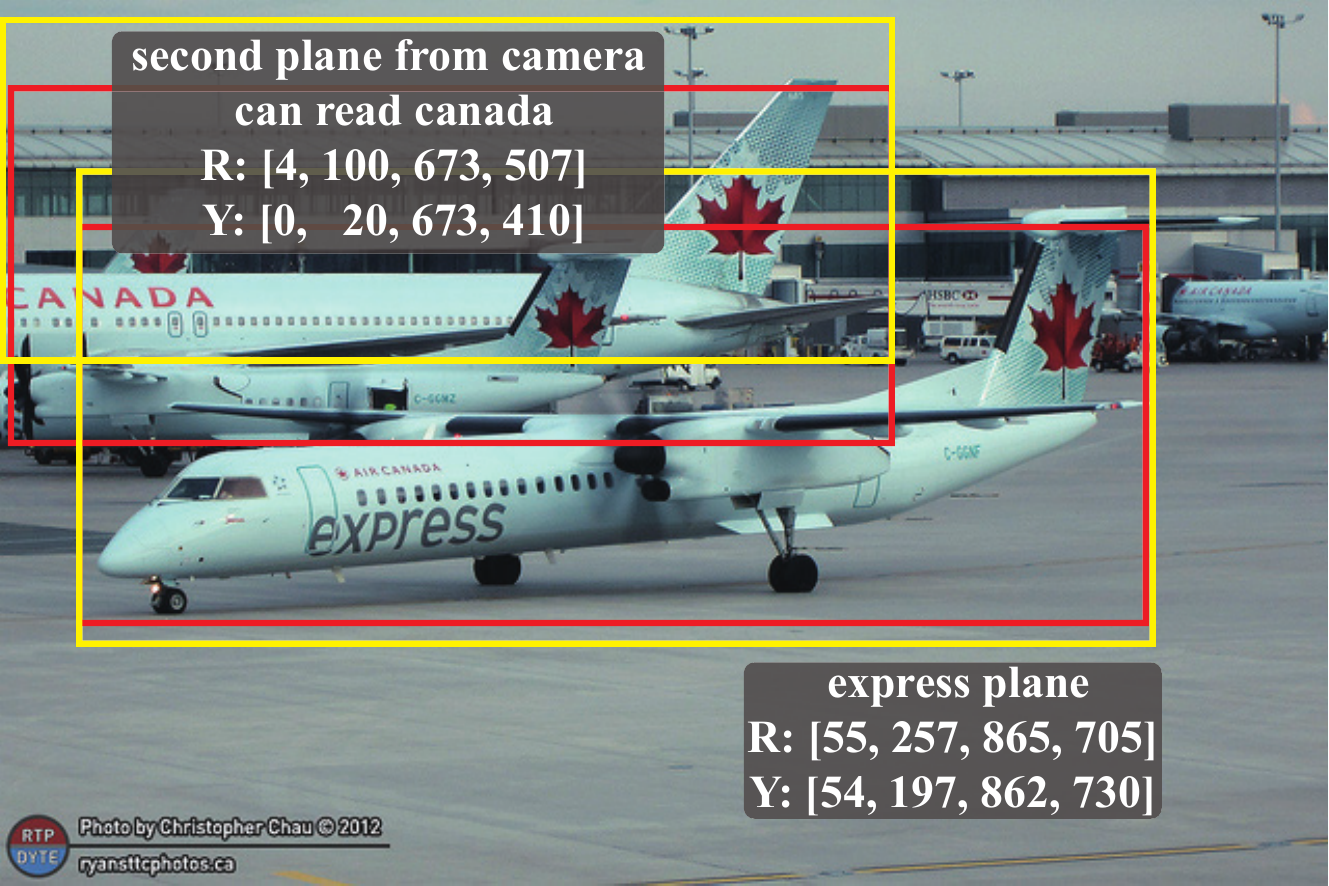}
        Non-person
        \caption{Referring Bbox Detection}
        \label{fig:casea}
    \end{subfigure}
    \begin{subfigure}[t]{0.33\textwidth}
        \centering
        \includegraphics[width=0.99\textwidth]{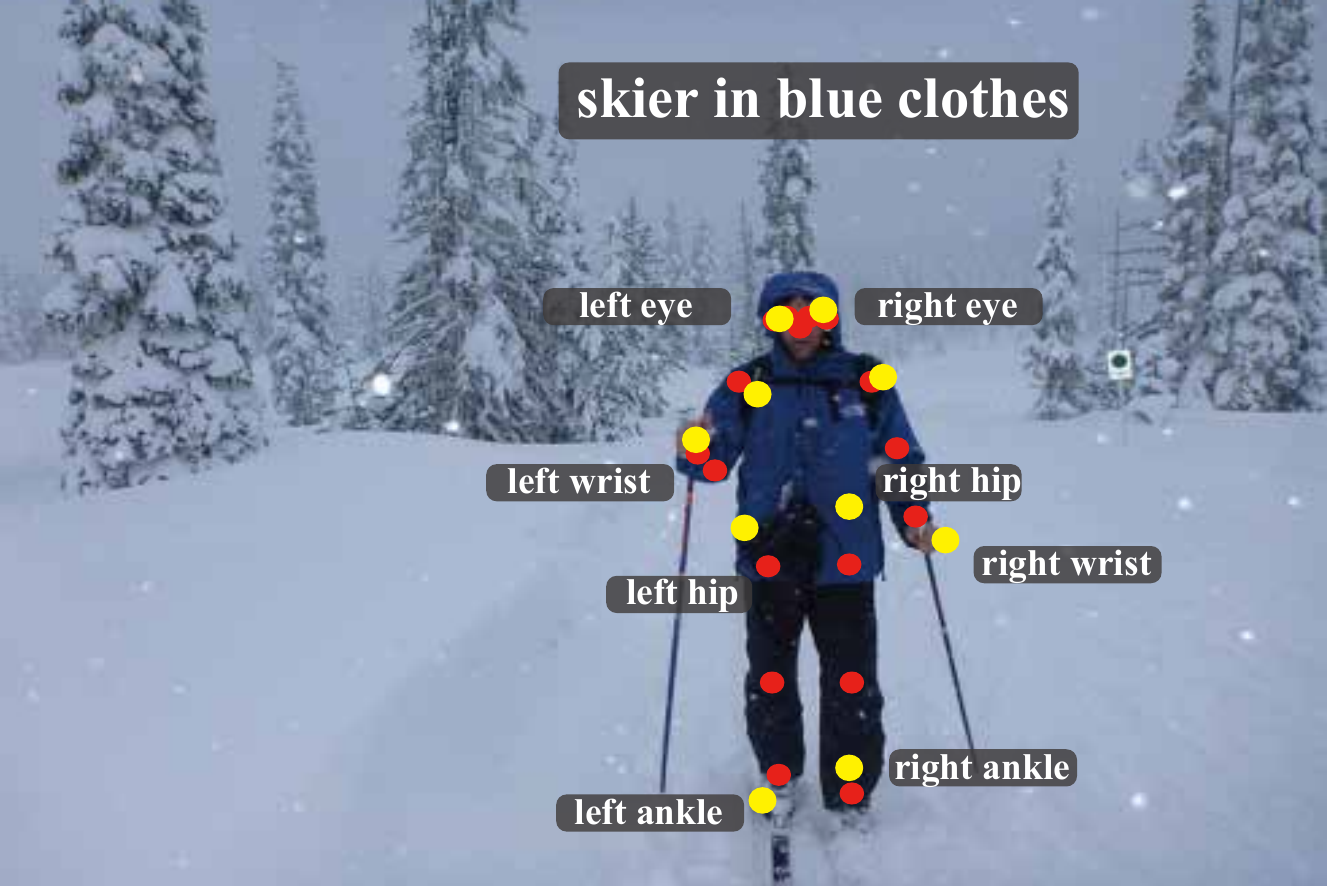}
        Person\\

        \includegraphics[width=0.99\textwidth]{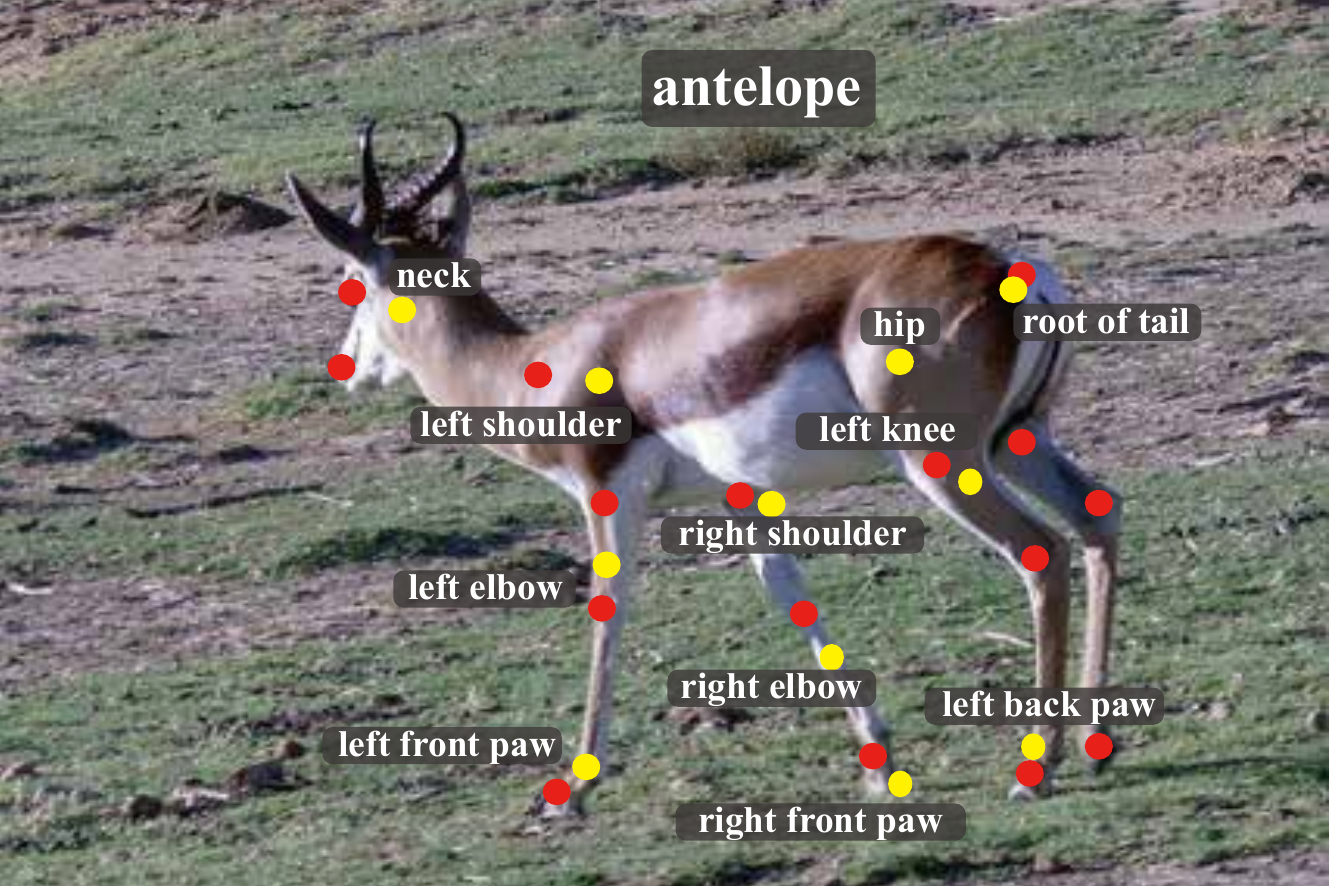}    
        Non-person
    
        \caption{Referring Keypoints Detection}
        \label{fig:caseb}
    \end{subfigure}
    \begin{subfigure}[t]{0.33\textwidth}
        \centering
        \includegraphics[width=0.99\textwidth]{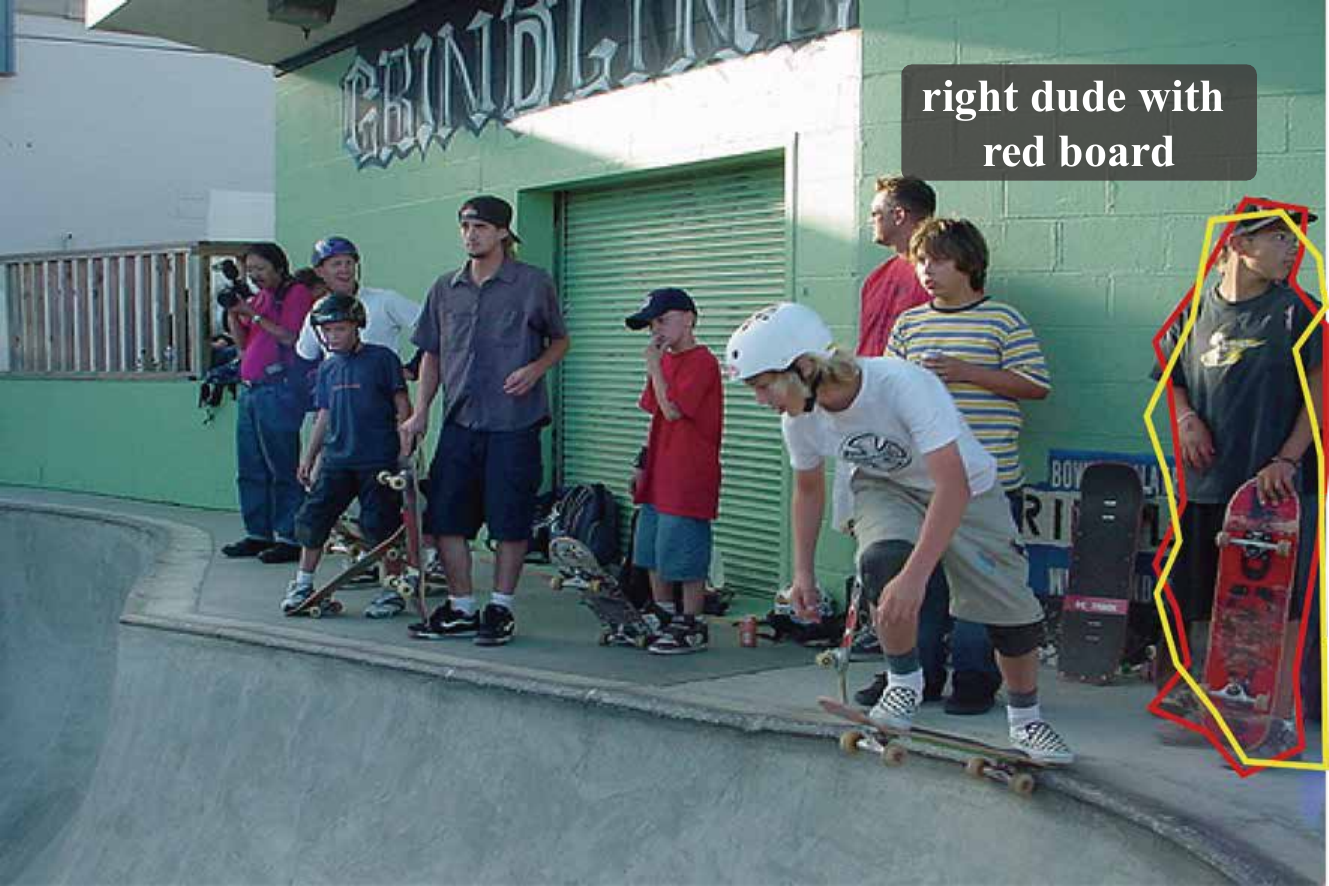}
        Persons\\

        \includegraphics[width=0.99\textwidth]{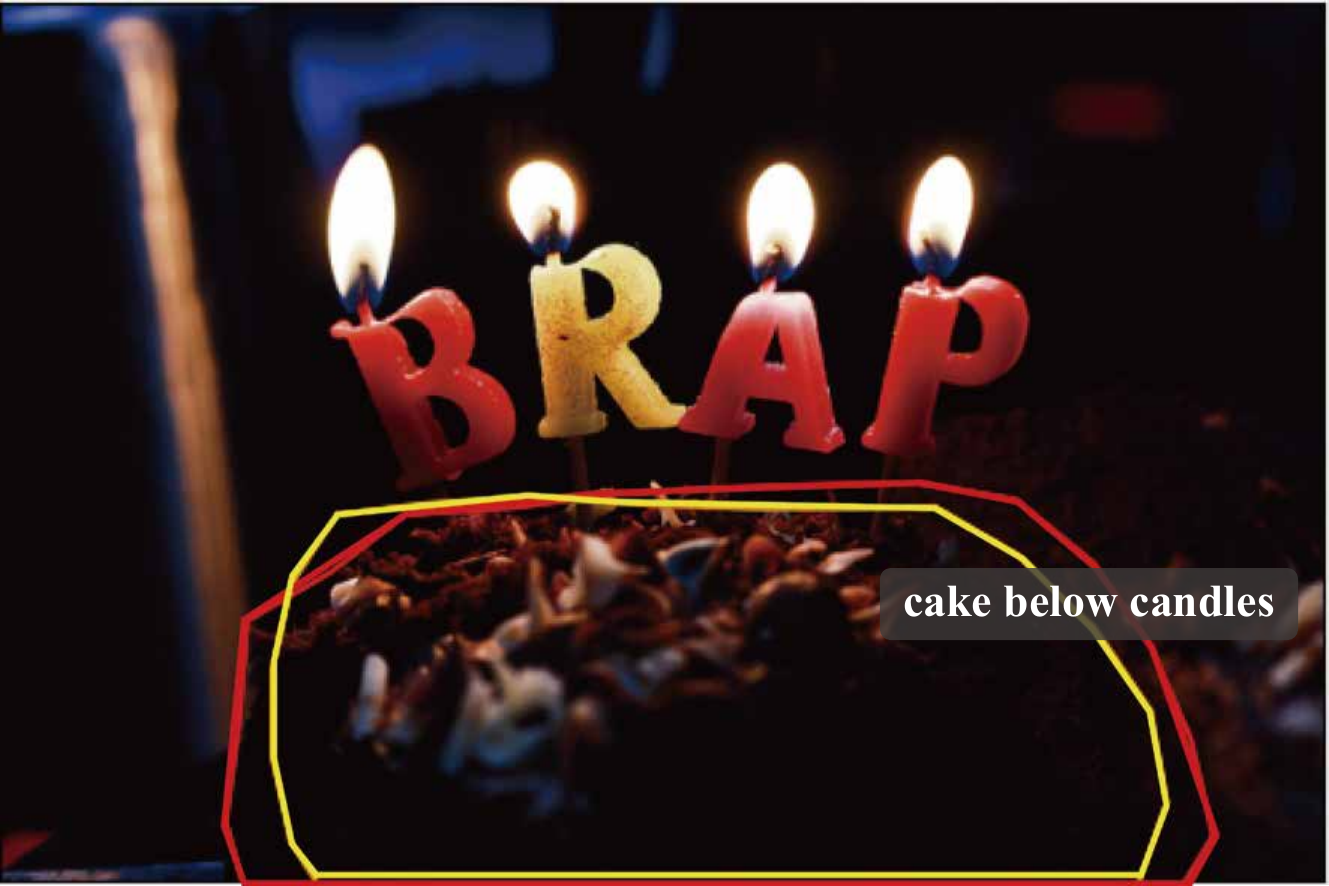}    
        Non-person
    
        \caption{Referring Image Segmentation}
        \label{fig:casec}
    \end{subfigure}   
    \caption{Results of multi-tasks of visual grounding, including (a) Referring bbox detection, (b) referring keypoints detection and (c) referring image segmentation. The red is represented as the ground truth and the yellow denotes the prediction. We choose the person and non-person to display the performance of our model in different tasks. }
    \label{fig:good-cases}
\end{figure*}  

\begin{figure}[htbp]
    \centering
    \begin{subfigure}[t]{0.49\textwidth}
        \centering
        \includegraphics[width=0.99\textwidth, height=0.14\textheight]{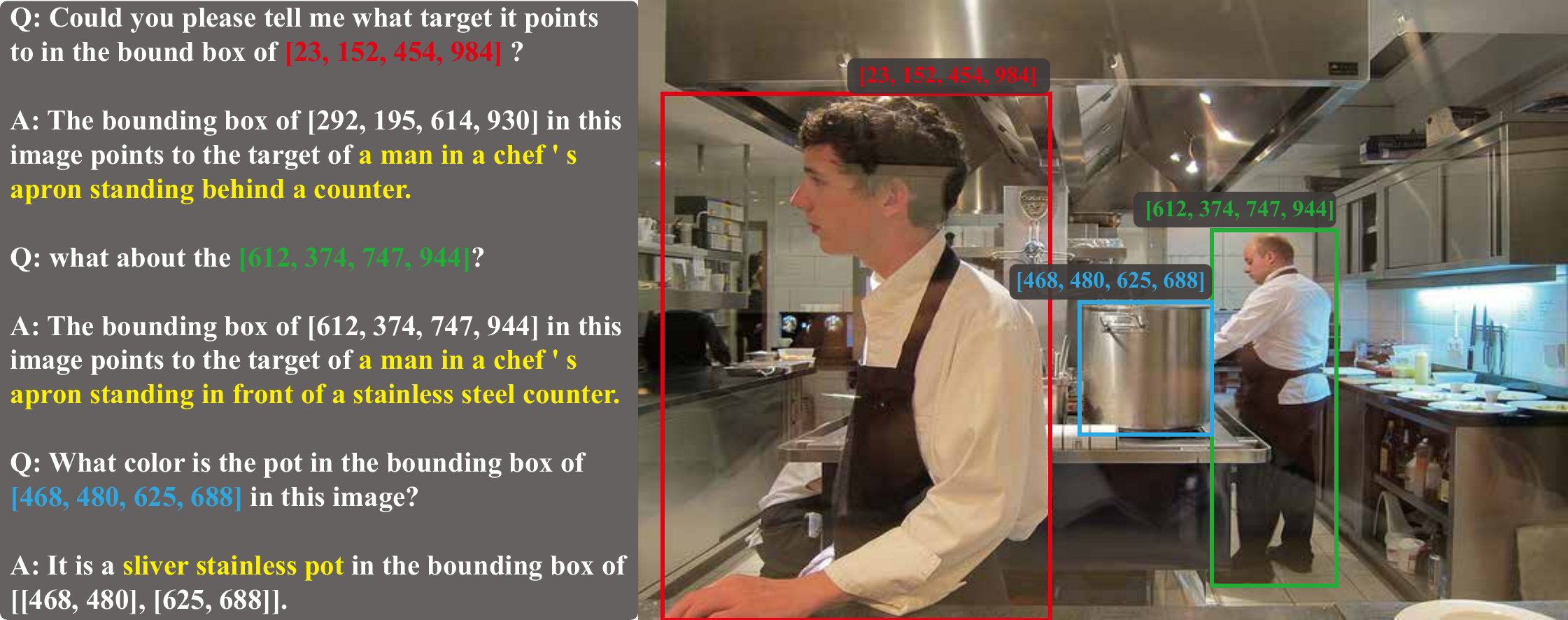}
        \caption{Referring caption based on the bbox}
        \label{fig:cased}
    \end{subfigure}
    \begin{subfigure}[t]{0.49\textwidth}
        \centering
        \includegraphics[width=0.99\textwidth, height=0.14\textheight]{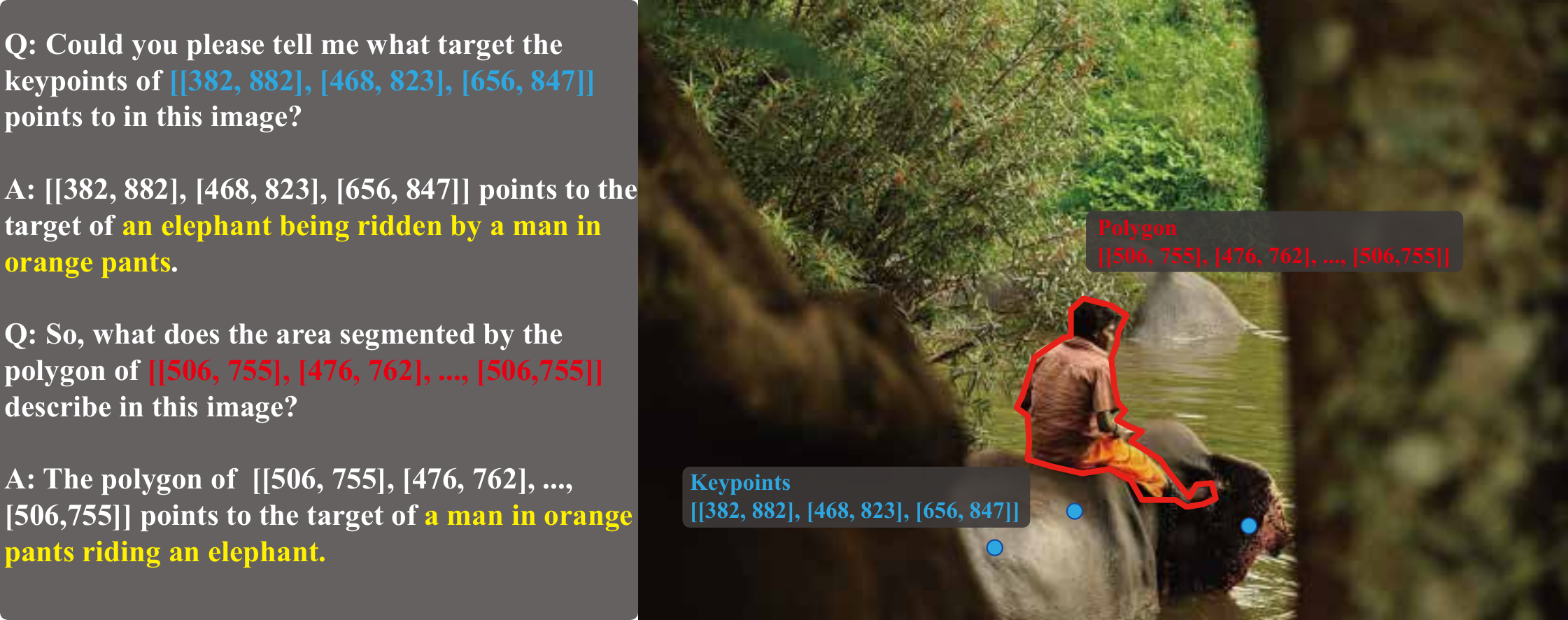}    
        \caption{Referring caption based on the keypoints and segmented polygon}
        \label{fig:casee}
    \end{subfigure}   
    \caption{Results of referring region caption}
    \label{fig:region-caption}
\end{figure}

\section{Methodology}

In this section, we first introduce the architecture of our vision-language model, ViLaM. Then, we explore the activation of object coordinates. Based on robustly outputting these object coordinates, we present the cycle training approach, which serves to reinforce the consistency between visual locations and their corresponding referring expressions.
Moreover, the cycle training also allows datasets without referring expressions to participate in the training, thereby facilitating the joint training of multiple tasks. Finally, we introduce our built VGcoco dataset, which includes multi-task annotations for visual grounding.

\subsection{Architecture}

\textbf{Image encoder: } With an input image $ x_{i}  \in  \mathbb{R}^{H\times W} $, visual features are extracted by image encoder and further projected to feature dimension:
\begin{equation}
    v_{i} = P_{img}(E_{img}(x_{i})) \in \mathbb{R}^{(h_{f} \times w_{f}) \times d}
\end{equation}
where $h_{f}$ and $w_{f}$ are the output size of visual features, and $d$ represents the feature dimension. $E_{img}$ can be any common visual backbones and we use Vit-Large in our case. Then by using $P_{img}$, which is composed of two linear layers, visual features are projected to feature dimension.

\noindent\textbf{Language encoder: } With any processed input instruction sequence $t_{i}$, text features are extracted by language encoder:
\begin{equation}
    l_{i} = E_{txt}(t_{i}) \in \mathbb{R}^{n_{t} \times d}
\end{equation}
where $n_{t}$ is the number of input tokens and $d$ represents the feature dimension. In our case, Bert \cite{devlin2019bert} is used as the language encoder.

\noindent\textbf{Multi-modality module: } This module follows an encoder-decoder architecture format. Given the input visual features $v_{i}$ and text features $l_{i}$, we first generate fused multi-modality representations by combining the image and text embeddings. These fused features serve as the keys and values in the cross-attention blocks in the decoder. By conditioning on the partial sequence $y_{i<j}$ predicted so far, the decoder recursively makes predictions for the token at position $j$, effectively generating aligned descriptions across modalities. 


\begin{equation}
    y_{i, j} = D_{mm}(E_{mm}(concat(v_{i}, l_{i})), y_{i <j}) \in \mathbb{R}^{1 \times d}
\end{equation}
whereas, $D_{mm}$ and $E_{mm}$ denote the decoder and the encoder of the multi-modality module, respectively. 

\subsection{Activation of Coordinates}

Leveraging its emergent capabilities, the vision-language model exhibits remarkable versatility in scenarios and tasks related to orientation. Initially, the task of activating coordinates is transmuted into conventional object detection within the framework of the vision-language model, devoid of referring expressions. Subsequently, extensive object detection datasets, such as COCO 2017, are integrated into the activation procedure. The considerable quantity of data facilitates the vision-language model in producing coordinates with enhanced robustness and precision.

Subsequently, to reconcile the divergence between semantic and linguistic coordinates, we establish a linguistic representation of coordinates within the large language model: $[x_{1}, y_{1}, x_{2}, y_{2}]$. Here, $x$ denotes horizontal coordinates, while $y$ signifies longitude. The pair $(x_{1}, y_{1})$ designates the upper-left point, and $(x_{2}, y_{2})$ corresponds to the lower-right point.  All coordinates adopt relative positions, normalization to 1000, and rounding.

We employ the captioning task to prompt our model to output coordinates that express orientation, owing to its proven effectiveness in capturing information in knowledge-intensive scenarios \cite{salaberriaImageCaptioningEffective2023}. During training, we utilize the captioning format as follows:
\begin{center}
    \centering
    \parbox{8cm}{\emph{find the $<$object$>$ in the region of $[x_{1}, y_{1}, x_{2}, y_{2}]$.}}
\end{center}
Due to the absence of referring expressions, more than one coordinate may correspond to multiple objects in the image.  The captioning form has more feasibility and practicability for the vision-language model to establish links between orientation and linguistic coordinates robustly, without inferring in the prompt.

In the training of captioning, the proposed model is expected to output image captions containing object-related coordinates and compute the loss. For activation of coordinates, we optimize using cross-entropy loss: 
\begin{equation}
    \mathcal{L}_{\mathrm{ce}} = - \sum_{i=1}^{n}\sum_{j=1}^{|y|} \log P_{\theta}(y_{i, j} | y_{i, <j}, x_i, t_i)
\end{equation}
where $n$ is the batch size, $\theta$ represents the model parameters, $x_{i}$ represents the input image, $t_{i}$ stands for the input instruction, and $y_{i, j}$ denotes the output token at position $j$ for the $ith$ sample at each batch. 
We follow the training strategy of BLIP2, which only trains the alignment layer and freezes the pre-trained visual model and large language model.
To enhance the quality of generation during inference, we employ various decoding techniques, such as beam search.

\subsection{Cycle Training}

Obtaining the ability of localization in the vision-language model, we design the cycle training to align and enhance the consistency between visual localization and referring expressions, as shown in Fig.\ref{fig:model}. Inspired by Cycle-GAN \cite{zhuUnpairedImagetoimageTranslation2017}, the cycle training is expected to learn alignment relationships between two domains $X$ and $Y$ given training samples $\{x_{i}\}_{i=1}^{N}\in X$ and $\{y_{j}\}_{j=1}^{M}\in Y$ in the vision-language model, where $X$ denotes the visual grounding features and $Y$ represents the linguistic referring expression.

The cycle training consists of two processes: referring expression generation (REG) represented as $G: X \rightarrow Y $  and referring expression comprehension (REC) formulated as $F: Y \rightarrow X $. The form of VQA is utilized to organize data, wherein we pose questions involving visual coordinates, and exploit the answers with referring expressions obtained from REG to construct new questions for REC. The visual localization from the answer of cycled REC is expected to be the same as the original. Vice versa, we also perform the cycle from REC to REG.




\begin{table}[]
    \caption{Experimental settings of tasks, datasets and metrics. It is worth noting that COCO 2017 does not contain referring expressions. Therefore, we do not utilize reference expressions when activating coordinates. While, in the tasks of referring bbox detection (RBD), referring keypoints detection (RKD) and referring image segmentation (RIS), the cycle training strategy generates the referring expressions for COCO 2017, enabling its participation in the training.}
    \label{table:settings}
    \centering
    \begin{tabular}{@{}cccc@{}}
    \toprule
    \multirow{2}{*}{Tasks}                                                                                            & \multicolumn{2}{c}{Datasets}                                                                  & \multirow{2}{*}{Metrics}                                                                  \\
                                                                                                                      & Training                                       & Valiadation                                  &                                                                                           \\ \midrule
    \begin{tabular}[c]{@{}c@{}}Activation\\ of Coordinates\end{tabular}                                               & COCO 2017                              &                                              &                                                                                           \\ \midrule
    \multirow{2}{*}{\begin{tabular}[c]{@{}c@{}}RBD\\ \end{tabular}}                         & RefCOCO/+/g                            & \multirow{2}{*}{RefCOCO/+/g}                   & \multirow{2}{*}{Acc@0.5}                                                                  \\
                                                                                                                      & COCO 2017                               &                       &                                                                                           \\ \midrule
    \multirow{3}{*}{\begin{tabular}[c]{@{}c@{}}RKD\\ \end{tabular}}                    & \multirow{3}{*}{COCO 2017}             & COCO 2017                              & \multirow{3}{*}{\begin{tabular}[c]{@{}c@{}}AP\\\end{tabular}}                                                   \\
                                                                                                                      &                                                & HumanArt                                     &                                                                                           \\
                                                                                                                      &                                                & AP-10K                                       &                                                                                           \\ \midrule
    \multirow{2}{*}{\begin{tabular}[c]{@{}c@{}}RIS\\\end{tabular}}                           & RefCOCO                                & \multirow{2}{*}{RefCOCO } & \multirow{2}{*}{\begin{tabular}[c]{@{}c@{}}IoU\\\end{tabular}} \\
                                                                                                                      & COCO 2017                              &                                              &                                                                                           \\ \midrule
    \begin{tabular}[c]{@{}c@{}}RRCls\\\end{tabular}                              &                                                & COCO 2017                              & ACC / mAP                                                                                 \\ \midrule
    \multirow{2}{*}{\begin{tabular}[c]{@{}c@{}}RRCap\\\end{tabular}}            &                                                & RefCOCOg                       & METEOR                                                                                    \\
                                                                                                                      &                                                & Visual Grenome                               & CIDEr                                                                                     \\ \midrule
    \begin{tabular}[c]{@{}c@{}}Medical Foreign\\  Object Detection\end{tabular}                     &                                                & Object-CXR                                   & Acc@0.5                                                                                   \\ \midrule
    \multirow{3}{*}{\begin{tabular}[c]{@{}c@{}}Disease \\ Localization \end{tabular}} & \multirow{3}{*}{\begin{tabular}[c]{@{}c@{}}ChestXray14 \\ (20-shot) \end{tabular}} & ChestXray14                            & \multirow{3}{*}{Acc@0.5}                                                                  \\
                                                                                                                      &                                                & TBX11K                                       &                                                                                           \\
                                                                                                                      &                                                & RSNA                                         &                                                                                           \\ \bottomrule
    \end{tabular}
\end{table}

With the above form of VQA, the visual localization and referring expression are cycled training in the vision-language model. We argue that the learned alignment relationships should be cycle-consistent: for every visual localization $x$ belonging to domain $X$, the cycle-referring expression should possess the capability to restore $x$ to its nearby coordinates, indicated $x\rightarrow G(x) \rightarrow F(G(x)) \approx x$. Similarly, for each referring expression $y$ from domain $Y$, $y$ should be reduced to its original form, i.e.  $y\rightarrow F(y) \rightarrow G(F(y)) \approx y$. The cycle referring expression can be incentivized by the consistency loss:
\begin{equation}
    \begin{aligned}
        \mathcal{L}_{\mathrm{cyc}}(G, F) & = \mathcal{L}_{\mathrm{ce}}(F(G(x)), x) + \mathcal{L}_{\mathrm{ce}}(F(G(y)), y)
    \end{aligned}        
\end{equation}

Benefiting from the cycle training, more normal object detection datasets without referring expressions could be expanded for the visual grounding training, such as COCO 2017. 
REG generates referring expressions from bounding boxes in COCO 2017, and REC then inferences the bounding boxes from the generated referring expression. 
Thereby, we can generate big data with visual localization and referring expressions to drive the large vision-language model to achieve visual grounding.


\subsection{Multi-Tasks Joint Learning}

We integrate multiple visual grounding tasks for joint training. Firstly, two types of tasks are participated in the training, namely referring expression comprehension (REC) and referring expression generation (REG). They are key factors of the cycle training. 

Among REC, various expression forms of visual coordinate are supported in our model, including bounding boxes, keypoints and segmented polygons, with varying granularity from the region-level to the pixel-level. 
To leverage these different modalities, we conducted joint training of REC by combining tasks of referring bounding box detection (RBD), referring keypoints detection (RKD), and referring image segmentation (RIS).
The coarse-grained RBD provides the target position information to guide the more fine-grained RKD and RIS tasks. In turn, the keypoints can effectively assist the polygons in deforming and outlining the boundaries of the targets. Vice versa, the pixel-level polygons also contribute to the refinement of bounding box and keypoint coordinates.

While in REG, referring region classification (RRCls) aims to determine the fine-grained category of the located target, and referring region caption (RRCap) focuses on generating a comprehensive description of the pointed target, including its category, location, color, size, and relationship with the surroundings. Through the cycle training strategy, the consistency between description and visual features is enhanced by the multi-tasks joint training.

\subsection{Building VGcoco for Visual Grounding}

Due to the inherent challenge of obtaining comprehensive location information, including bounding boxes, keypoints, and segmented polygons, most visual grounding datasets struggle to offer location details at various levels of granularity.
Recognizing the demand for higher-quality referring expression data that encompasses diverse forms of visual location, we developed a dataset called VGCoco. It contains about 240K images with grounding varying from region-level to pixel-level and corresponding referring expressions. 

\begin{table*}[htbp]
    \caption{Evaluation results of referring bbox detection on RefCOCO, RefCOCO+ and RefCOCOg datasets. The best-performing multi-tasks models are highlighted in red, and the second-best in blue. Acc@0.5 is applied to evaluate the performance of different methods. 
    }
    \label{tab:vg-results}
    \centering
    \setlength{\tabcolsep}{1mm}{
    \begin{tabular}{@{}lccccccccccc@{}}
    \toprule
                                                                                       &    &&                                   & \multicolumn{3}{c}{RefCOCO}                                                                                           & \multicolumn{3}{c}{RefCOCO+}                                                                                          & \multicolumn{2}{c}{RefCOCOg}                                                  \\ \cmidrule(l){5-12} 
    \multirow{-2}{*}{Models}                                                           & \multirow{-2}{*}{Type} & \multirow{-2}{*}{Visual Encoder} & \multirow{-2}{*}{Language Model}              & val                                   & testA                                 & testB                                 & val                                   & testA                                 & testB                                 & val-u                                 & test-u                                \\ \midrule
     SeqTR                                   \cite{Zhu_2022}                            &                                      &Darknet-53&Bi-GRU & 83.72                                 & 86.51                                 & 81.24                                 & 71.45                                 & 76.26                                 & 64.88                                 & 74.86                                 & 74.21                                 \\
    MDETR                                   \cite{kamath2021mdetr}                     &                                       &EfficientNet-B3&RoBERTa-base & 86.75                                 & 89.58                                 & 81.41                                 & 79.52                                 & 84.09                                 & 70.62                                 & 81.64                                 & 80.89                                 \\
    VGTR                                    \cite{du2022visual}                        &  \multirow{-3}{*}{\begin{tabular}[c]{@{}c@{}}Specialized\\ Models\end{tabular}}                                    &EfficientNet-B3&RoBERTa-base  & 79.30                                 & 82.16                                 & 74.38                                 & 64.40                                 & 70.85                                 & 55.84                                     & 66.83                                 & 67.28      \\
    \midrule
    OFA                                     \cite{wang2022ofa}                         &                                      &ResNet-152&BART-Large & {\color[HTML]{0000FF} \textbf{92.04}} &{\color[HTML]{0000FF} \textbf{94.03}}                                 & {\color[HTML]{0000FF} \textbf{88.44}}                                 & {\color[HTML]{0000FF} \textbf{87.86}} & {\color[HTML]{0000FF} \textbf{91.70}}                                 & {\color[HTML]{0000FF} \textbf{80.71}}                                 & {\color[HTML]{0000FF} \textbf{88.07}} & {\color[HTML]{0000FF} \textbf{88.78}} \\
    mPLUG-2                                 \cite{xuMPLUG2ModularizedMultimodal2023}   &                                       &ViT-L/14&BERT-Large& 90.33                                 & 92.80                                 & 86.05                                 & -                                     & -                                     & -                                     & 84.70                                 & 85.14                                 \\
    FERRET                              \cite{youFerretReferGround2023}                      &                              &ViT-L/14&  Vicuna-7B      & 87.49                                 & 91.35                                 & 82.45                                 & 80.78                                 & 87.38                                 & 73.14                                 & 83.93                                 & 84.76                                 \\
    \textbf{Ours}                                                                      & \multirow{-4}{*}{\begin{tabular}[c]{@{}c@{}}Generalist\\ Models\end{tabular}}  &ViT-L/14&Vicuna-7B& {\color[HTML]{FF0000} \textbf{92.99}} & {\color[HTML]{FF0000} \textbf{95.90}} & {\color[HTML]{FF0000} \textbf{90.39}} & {\color[HTML]{FF0000} \textbf{90.96}}   & {\color[HTML]{FF0000} \textbf{94.78}} & {\color[HTML]{FF0000} \textbf{86.93}} & {\color[HTML]{FF0000} \textbf{90.05}} & {\color[HTML]{FF0000} \textbf{89.51}} \\ \bottomrule
    \end{tabular}
    }
\end{table*}

We extend the open-source datasets, namely COCO-Pose \cite{lin2015microsoft}, AP-10K \cite{yu2021ap} and a part of COCO 2017 \cite{linMicrosoftCOCOCommon2014}. With the aid of the cycle training strategy, the reference expressions required by COCO-Pose are generated by our model. Similarly, AP-10 only contains the keypoints of animal skeletons and bounding boxes, so the segmented polygons and referring expressions are produced by our ViLaM. In addition to persons at COCO-Pose and animals at AP-10K, a part of common objects at COCO 2017 are joined in the designed datasets. We use skeletonization and clustering methods to obtain the keypoints of the target, and generate its reference expression through our model. 
To support and encourage the community focused on visual grounding, we have released this dataset to the public. For further information about VGcoco, please refer to the supplementary materials.

\section{Experiments}

\subsection{Experimental Settings}

The experimental settings are exhibited in the Table.\ref{table:settings}, including tasks, dataset and metrics. Our training dataset primarily contains the 
COCO 2017 \cite{linMicrosoftCOCOCommon2014},
RefCOCO \cite{yuModelingContextReferring2016}, 
RefCOCO+ \cite{yuModelingContextReferring2016}, 
and RefCOCOg \cite{maoGenerationComprehensionUnambiguous2016, nagarajaModelingContextObjects2016}. 
Besides, we use the ChestXray14 \cite{wang2017chestx} as the 20-shot training dataset to perform a case study of generalization of disease localization on chest X-rays.

For validation, in addition to assessing the accuracy of our model on the respective closed-set dataset, we also evaluated its generalizability in an open-set scenario, such as HumanArt \cite{juHumanArtVersatileHumanCentric2023} and AP-10K \cite{yu2021ap} in referring keypoints detection, Visual Grenome \cite{krishnaVisualGenomeConnecting2017} in referring region caption. Besides, the generalization is validated from different domains, such as Object-CXR \cite{objectcxr} in medical foreign object detection and RSNA Pneumonia dataset \cite{shih2019augmenting}, TBX11K dataset \cite{liu2020tbx11k} and ChestXray14\cite{wang2017chestx} in disease localization on Chest X-ray.

When assessing the performance of our model across multiple tasks, the Acc@0.5 evaluates the impact of bbox-related tasks, OKS-based AP validates the performance of referring keypoints detection, and IoU measures the effectiveness of referring image segmentation. Furthermore, in order to evaluate the capability of referring region caption, we select the task of classification to output the category 
in the bbox with ACC metric
and the task of the detailed caption to generate the description with metrics of METEOR and CIDEr.
More detailed experimental implementations are given in the supplementary.

\subsection{Accurate REC of Multi-Tasks}

Qualitatively, Fig.\ref{fig:good-cases} exhibits the visual grounding results of multi-tasks, including referring bbox detection, referring keypoints detection and referring image segmentation. For each task, we exhibit the performance of visual grounding in person and non-person. The red denotes the ground truth and the yellow is the prediction. 

\begin{table}[htbp]
    \caption{Evaluation results of referring keypoints detection on the close-set scenario with the COCO val2017 dataset, and open-set generalization validation over the HumanArt and AP-10K datasets. 
    The best-performing multi-tasks models are highlighted in red, and the second-best in blue. Average precision is utilized to validate the performance of keypoints detections.}
    \label{table:keypoints}
    \centering
    \begin{tabular}{@{}lcccc@{}}
        \toprule
        Models                                                               & Type                                                                      & COCO val                              & HumanArt                              & AP-10K                                  \\ \midrule
        PCT          \cite{gengHumanPoseCompositional2023}              &                                                                                & {\color[HTML]{A6A6A6} 80.20}          & {\color[HTML]{A6A6A6} 63.70}          & {\color[HTML]{A6A6A6} 14.60}            \\
        ViTPose      \cite{xuViTPoseSimpleVision2022}                   & \multirow{-2}{*}{\begin{tabular}[c]{@{}c@{}}Specialized\\ Models\end{tabular}} & {\color[HTML]{A6A6A6} 82.00}          & {\color[HTML]{A6A6A6} 64.10}          & {\color[HTML]{A6A6A6} 14.70}            \\ \midrule
        Unified-IO   \cite{luUNIFIEDIOUnifiedModel2022}                 &                                                                                & 25.00                                 & 15.70                                 & 7.60                                    \\
        Painter      \cite{wangImagesSpeakImages2023}                   &                                                                                & 70.20                                 & 12.40                                 & 15.30                                   \\
        InstructDiff \cite{gengInstructDiffusionGeneralistModeling2023} &                                                                                & {\color[HTML]{0000FF} \textbf{71.20}} & {\color[HTML]{0000FF} \textbf{51.40}} & {\color[HTML]{0000FF} \textbf{15.90}}   \\
        $\textbf{Ours}$                                                 & \multirow{-4}{*}{\begin{tabular}[c]{@{}c@{}}Generalist\\ Models\end{tabular}}  & {\color[HTML]{FF0000} \textbf{76.10}} & {\color[HTML]{FF0000} \textbf{54.62}} & {\color[HTML]{FF0000} \textbf{44.67}}   \\ \bottomrule
    \end{tabular}

\end{table}

\begin{table}[htbp]
    \caption{Evaluation results of referring image segmentation on RefCOCO dataset. The best-performing generalist models are highlighted in red, and the second-best in blue. IoU is utilized to validate the performance of segmentation.}
    \label{table:polygon}
    \centering
    \begin{tabular}{@{}lcccc@{}}
        \toprule
        \multicolumn{1}{c}{}                                                &                                                                                & \multicolumn{3}{c}{RefCOCO}                                                                                               \\ \cmidrule(l){3-5} 
        \multicolumn{1}{c}{\multirow{-2}{*}{Methods}}                       & \multirow{-2}{*}{Type}                                                         & val                                   & testA                                  & testB                                    \\ \midrule
        LAVT           \cite{yangLAVTLanguageAwareVision2022}               &                                                                                & {\color[HTML]{A6A6A6} 74.46}          & {\color[HTML]{A6A6A6} 76.89}           & {\color[HTML]{A6A6A6} 70.94}             \\
         SeqTR          \cite{Zhu_2022}                                      &                                                                                & {\color[HTML]{A6A6A6} 71.70} & {\color[HTML]{A6A6A6} 73.31}  & {\color[HTML]{A6A6A6} 69.82}    \\
        PolyFormer     \cite{liuPolyFormerReferringImage2023a}              & \multirow{-3}{*}{\begin{tabular}[c]{@{}c@{}}Specialized\\ Models\end{tabular}} & {\color[HTML]{A6A6A6} 74.82}          & {\color[HTML]{A6A6A6} 76.64}           & {\color[HTML]{A6A6A6} 71.06}             \\ \midrule
        Unified-IO     \cite{luUNIFIEDIOUnifiedModel2022}                   &                                                                                & 46.42                                 & 46.06                                  & 48.05                                    \\
        InstructDiff   \cite{gengInstructDiffusionGeneralistModeling2023}   &                                                                                & 61.74                                 & 65.20                                  & 60.17                                    \\
        LISA          \cite{lai2023lisa}                                      &                                                                                & {\color[HTML]{0000FF} \textbf{74.10}} & {\color[HTML]{FF0000} \textbf{76.50}}  & {\color[HTML]{0000FF} \textbf{71.10}}    \\
        $\textbf{Ours}$                                                     & \multirow{-4}{*}{\begin{tabular}[c]{@{}c@{}}Generalist\\ Models\end{tabular}}  & {\color[HTML]{FF0000} \textbf{74.85}}      & {\color[HTML]{0000FF} \textbf{76.02}}  & {\color[HTML]{FF0000} \textbf{74.34}}    \\ \bottomrule
    \end{tabular}

\end{table}

For referring bbox detection, Fig.\ref{fig:casea} illustrates the superiority of our method. It accurately identifies the object and understands its description words, such as position, color, and text on the object. Significantly, our model demonstrates the ability to recognize objects that are overlapping and occluded. When overlapping targets of the same class are present, our model reveals remarkable capabilities in understanding, discriminating, and locating them. This further corroborates that the cycle training effectively enhances the consistency between visual location and referring expressions.

In the context of referring keypoints detection illustrated in Fig.\ref{fig:caseb}, our model leverages the power of large language models to proficiently identify the corresponding keypoints by employing simple interactive questions, such as eyes, hips and shoulders. In joint training of multi-tasks, the referring keypoints detection empowers the model to gain a deeper understanding of the relationships among various targets. In addition to the referring expression of keypoints-level, we extend the target-level referring expression through the cycle training strategy, such as changing "person" to "skier in blue clothes". It enables the integration of target-level referring expressions during the joint training, thereby enhancing the model's comprehension capabilities.

Besides, our model expands the competencies of visual grounding to pixel-level shown in Fig.\ref{fig:casec}. Polygon is adopted to accurately delineate the objects denoted by referring expressions,  thereby significantly enhancing the alignment between the target shape and the referring expressions through the cycle training strategy.

Quantitatively, Table.\ref{tab:vg-results} presents a comparison of referring bbox detection results between our model and various types of visual grounding models, including specialized models and multi-tasks models. It clearly demonstrates that our method achieves SOTA performance across all test datasets. Notably, in the testB split of RefCOCO and RefCOCO+, our model outperforms other methods by a significant margin. This highlights the superiority of our approach in effectively handling the referring expression comprehension with bounding boxes, particularly for non-people objects.

Moreover, Table.\ref{table:keypoints} indicates our model performs excellently in the referring keypoints detection, with the best performance among multi-tasks models. Notably, our model demonstrates significantly stronger generalization performance on the open-set datasets of HumanArt and AP-10K compared to other models. In fact, we surpass specialized models by a significant margin in AP-10K, corroborating our main contribution of the cycle training strategy, which enhances the capability of LLMs in understanding, discriminating, locating and generalization.


Additionally, Table.\ref{table:polygon} investigates the effectiveness of our model in referring image segmentation. Our performance surpasses other methods, with the exception of being slightly inferior to LISA\cite{lai2023lisa} on the testA split of RefCOCO dataset, which employs ViT-H SAM \cite{kirillov2023segment} backbone as the vision backbone.

        
\begin{table}[htbp]
    \caption{Evaluation results of referring object classification on COCO 2017 val set. The best-performing generalist models are highlighted in red, and the second-best in blue. ACC is utilized to validate the performance of classification.}
    \label{table:classification}
    \centering
    \begin{tabular}{@{}lccccc@{}}
        \toprule
        & & &\multicolumn{3}{c}{Ours}\\ \cmidrule(l){4-6} 
\multirow{-2}{*}{LLaVA \cite{liu2023improved}}  
& \multirow{-2}{*}{Shikra\cite{chen2023shikra} }
& \multirow{-2}{*}{PVIT  \cite{chen2023position}}
& keypoints & bbox & polygon  \\ \midrule       
        40.04 &53.91&64.53       & 74.52 &{\color[HTML]{0000FF} \textbf{80.58}}&   {\color[HTML]{FF0000} \textbf{81.55}}     \\ \bottomrule
    \end{tabular}                        
\end{table}

\begin{table}[htbp]
    \caption{Evaluation results of the referring region caption on RefCOCOg and Visual Genome dataset. The best-performing multi-tasks models are highlighted in red, and the second-best in blue. METEOR and CIDEr is utilized to validate the performance of region captioning.}
    \label{table:caption}
    \centering
    \begin{tabular}{@{}lccccc@{}}
        \toprule
        \multicolumn{1}{c}{}                                                &   \multicolumn{2}{c}{RefCOCOg}                                                                              & \multicolumn{2}{c}{Visual Genome}                                                                                               \\ \cmidrule(l){2-5} 
        \multicolumn{1}{c}{\multirow{-2}{*}{Methods}}                      
        & METEOR                                   & CIDEr                                  & METEOR& CIDEr                                     \\ \midrule
GRIT \cite{wu2022grit} &15.2& 71.6 & {\color[HTML]{000000} \textbf{17.1}}&{\color[HTML]{FF0000} \textbf{142.0}}    \\
        Kosmos-2     \cite{peng2023kosmos}     &  14.1                       &62.3      &-&-     \\
        $\textbf{Ours(keypoints)}$  & {\color[HTML]{0000FF} \textbf{26.5}} & 146.6 & 17.0 & {\color[HTML]{000000} \textbf{130.1}}\\
        $\textbf{Ours(bbox)}$  & 26.3 & {\color[HTML]{FF0000} \textbf{167.2}} & {\color[HTML]{FF0000} \textbf{19.7}} & {\color[HTML]{0000FF} \textbf{131.3}}\\ 
        $\textbf{Ours(polygon)}$  & {\color[HTML]{FF0000} \textbf{26.6}} & {\color[HTML]{0000FF} \textbf{165.6}} & {\color[HTML]{0000FF} \textbf{18.8}}&105.9\\
        \bottomrule
    \end{tabular}
                                   
\end{table}

\subsection{Region Captioning}

\noindent\textbf{Referring Object Classification}
The performance of object classification was evaluated on COCO 2017 dataset using classification accuracy. As shown in Table.\ref{table:classification}, the referring object classification task achieved excellent results using different visual prompts,i.e., keypoints, bbox, and polygon. Compared to PVIT method, there was an approximately 15\% improvement in the ACC values with the bbox prompt of our method. We observed a significant improvement when using the polygon prompt, indicating an inherent relationship between the shape and the object.

\noindent\textbf{Referring Object Caption}
We further evaluate the region-level captioning ability of our model on the RefCOCOg and Visual Genome datasets. As shown in Table.\ref{table:caption}, the region-level caption task has also achieved excellent results on the RefCOCOg dataset, with improvements of more than 10\% and 90\% in METEOR and CIDEr scores, respectively, compared to the results achieved by the Kosmos-2 method. There is no obvious difference among the results obtained by various visual prompts.
More ablation experiments for the caption task are provided in the supplementary materials.

\begin{figure}[htbp]
    \centering
    \includegraphics[width=0.48\textwidth, height=0.3\textheight]{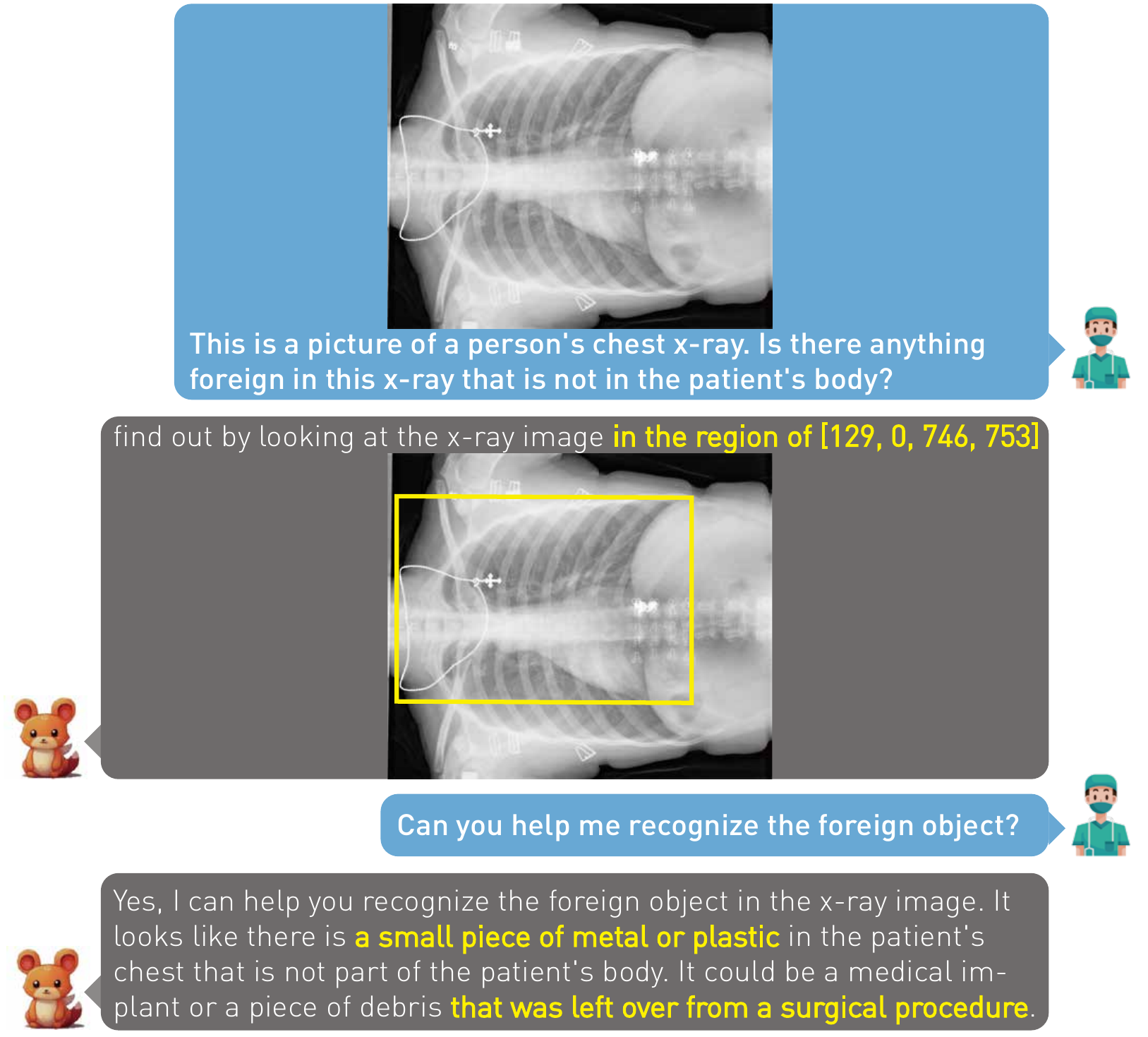}
    \caption{The zero-shot results of visual question answering for foreign objects detection in chest X-ray images. 
    The presence of foreign objects is accurately detected by giving its coordinates. Particularly, the model can deduce that the foreign object is metal or plastic by asking to recognize the foreign object. }
    \label{fig:foreign}
\end{figure}

\subsection{Case Studies of Robust Generalization in the Medical Domain}

We conduct several case studies in the medical domain to corroborate the robust generalization of our model, namely medical foreign object detection and disease localization in chest X-rays.

Regarding medical foreign object detection, we employed the model trained on natural images directly, without fine-tuning specifically on chest X-rays. 
Fig.\ref{fig:foreign} illustrates the results of visual question and answering for medical foreign object detection on the Object-CXR dataset \cite{objectcxr}. The necklace around the neck and the clip in the middle retain their inherent shape in X-rays, so our model accurately identifies the presence of these foreign objects outside the body, attributed to its robust and considerable generalization. Beyond the localization, our model can recognize the detected foreign object upon inquiry, and deduce that the object is likely made of metal or plastic debris. More specially, ViLaM also exhibits an ability to infer the potential origin or source of the debris, leveraging its extensive language understanding capacity. Moreover, on the quantitative of classify the presence or absence of foreign objects, ViLaM achieves an AUC of 93.1\%, demonstrating a substantial feasibility and robust generalization.

\begin{table}[htbp]
\caption{Evaluation results of \textbf{disease localization} task with 20-shot setting on four typical disease labels from the chest X-ray datasets. Acc@0.5 is applied to evaluate methods.}
    \centering
    \begin{tabular}{@{}lcccc@{}}
    \toprule
    Datasets & TBX11K       & RSNA      & \multicolumn{2}{c}{ChestXray14}                       \\
    Diseases & Tuberculosis & Pneumonia & Atelectasis & Pneumothorax \\ \midrule
    VGTR \cite{du2022visual}    & 1.99       & 4.67    & 3.70          & 0       \\ 
    OFA    \cite{wang2022ofa}  & 20.40       & 14.67    & 3.90        & 12.49       \\
    Ours  & 30.84       & 28.00    & 11.11        & 20.83       \\ 
     \bottomrule
    \end{tabular}

    \label{tab:diseaselocation}
\end{table}

To further examine the generalization and scalability of our model in the medical field,
we conduct preliminary experiments of disease localization on three typical chest X-ray datasets, namely, TBX11K\cite{liu2020tbx11k}, RSNA Pneumonia\cite{shih2019augmenting}, and ChestXray14\cite{wang2017chestx}.
The proposed model is fine-tuned with 20-shot labels for each disease of ChestXray14 datasets.
As depicted in Table.~\ref{tab:diseaselocation}, ViLaM consistently outperforms other approaches in various disease categories. Particularly noteworthy is the significant improvement of more than 15\% in detecting Pneumonia and Pneumothorax compared to alternative methods. This robust performance further validates the exceptional generalization and practicality capability of our model.
More experiment results are available in the supplement material.

\subsection{Ablation Study}

\subsubsection{Cycle Training}

\begin{table}[htbp]
    \caption{Ablation evaluation results of referring bbox detection on RefCOCO dataset with different modules. \checkmark denotes the applied module. Acc@0.5 is utilized to evaluate the performance of various conditions.}
    \label{table:ablation}
    \centering
    \begin{tabular}{@{}cccccc@{}}
    \toprule
    \multicolumn{3}{c}{Modules}                                                                                                                                                          & \multicolumn{3}{c}{RefCOCO} \\ \midrule
    \begin{tabular}[c]{@{}c@{}}Coordinates \\ Activation\end{tabular} & \begin{tabular}[c]{@{}c@{}}Cycle \\ Training\end{tabular} & \begin{tabular}[c]{@{}c@{}}Cycle \\ Augment.\end{tabular}           & val     & testA   & testB   \\ \midrule
    \checkmark                                                           &                                                           &                                                              & 79.95   & 79.24   & 80.99   \\
    \checkmark                                                           & \checkmark                                                &                                                              & 85.59   & 87.54   & 82.60   \\
    \checkmark                                                           & \checkmark                                                & \checkmark                                                   & 92.99   & 95.90   & 90.39   \\ \bottomrule
    \end{tabular}
\end{table}

Firstly, we conduct ablation studies to verify the improvement of the cycle training strategy. The cycle training brings about improvements through two key factors. First, it enhances the consistency between reference expressions and their related locations. Second, additional datasets without referring expressions can participate in the training by generating referring expressions through cycle training. 

Consequently, we conduct ablation experiments to evaluate the impact brought by these two key factors of cycle training in Table.\ref{table:ablation}, where "Cycle Training" denotes the performance gain from enhancing consistency using cycle training, and "Cycle Augment." represents the boost from additional datasets, which uses cycle training to generate referring expressions. The referring bbox detection is employed as the task of validation using RefCOCO dataset.
Table.\ref{table:ablation} exhibits that the improvement from the enhanced consistency achieves an Acc@0.5 of 5.64\% in val, 8.3\% in testA and 1.61\%, and from the additional datasets is Acc@0.5 of 7.4\% in val, 8.36\% in testA and 7.79\%.

Upon analysis, we observe that the testA split of RefCOCO exclusively comprises people, thereby leading to a more pronounced improvement resulting from the enhanced consistency achieved through cycle training, as compared to the testB split, which includes non-people. Besides, data augmentation exhibits a significant enhancement effect in different splits.

\begin{table}[]
\caption{Ablation evaluation results of multi-tasks joint training of visual grounding. \checkmark denotes the participated tasks. RIS and RKD are exploited as the validation tasks, and IoU and AP are opted for the metrics, respectively.}
\label{table:abl-joint}
\centering
\begin{tabular}{@{}ccccccc@{}}
\toprule
\multicolumn{3}{c}{Multi-Task Joint Training}                                                                                                                                                                                                                            & \multicolumn{3}{c}{\begin{tabular}[c]{@{}c@{}}RIS(IoU)\\ \end{tabular}} & \begin{tabular}[c]{@{}c@{}}RKD(AP)\\\end{tabular} \\ \midrule
\multirow{2}{*}{\begin{tabular}[c]{@{}c@{}}RBD\\ \end{tabular}} & \multirow{2}{*}{\begin{tabular}[c]{@{}c@{}}RKD\\ \end{tabular}} & \multirow{2}{*}{\begin{tabular}[c]{@{}c@{}}RIS \\ \end{tabular}} & \multicolumn{3}{c}{RefCOCO}                                                                      & \multirow{2}{*}{COCO val}                                                    \\
                                                                                    &                                                                                          &                                                                                          & val                            & testA                          & testB                          &                                                                              \\ \midrule
\checkmark                                                           & \checkmark                                                                                        &                                                                                          & -                              & -                              & -                              & 70.46                                                                        \\
\checkmark                                                                                   &                                                                                          & \checkmark                                                                                        & 62.83                          & 63.28                          & 61.06                          & -                                                                            \\
\checkmark                                                                                   & \checkmark                                                                                        & \checkmark                                                                                        &    74.85                            & 76.02                          & 74.34                          & 76.10                                                                         \\ \bottomrule
\end{tabular}
\end{table}

\subsubsection{Joint Training}

Furthermore, we implement ablation experiments to validate the effect of multi-tasks joint training of visual grounding in Table.\ref{table:abl-joint}. The referring keypoints detection and referring image segmentation are selected as the tasks of validation. Since the bbox detection is utilized to activate the coordinates of LLMs, the task of referring bbox detection is included in all ablation experiments.

The results of Table.\ref{table:abl-joint} showcases the joint learning of multi-tasks of visual grounding substantially boosts the performance of individual tasks. In particular, the referring keypoints detection considerably improves the segmentation task. We analyze that it is mainly because keypoints can well assist polygons in deforming and outlining the boundaries of the targets.

\section{Conclusion}
We have developed a vision-language model, ViLaM, that enhances visual grounding capabilities and generalization performance based on the foundations of a large language model. Despite being trained solely on the COCO 2017 and RefCOCO/+/g datasets, we are able to generate a considerable amount of additional annotations through the cycle training for multi-tasks, and exhibit competitive performance on multiple referring expression comprehension and generation tasks. 
We further contribute a multi-task dataset, encompassing referring expression and related annotations of bounding boxes, keypoints, and segmented polygons.

\bibliographystyle{unsrt}  
\bibliography{references}  

\section*{Appendix}

\begin{table}[htbp]
  \caption{Overview of  12 medical datasets across 6 modalities.}
  \label{table:datasets}
  \centering
  \setlength{\tabcolsep}{3mm}{
  \begin{tabular}{@{}lccc@{}}
  \toprule
  \begin{tabular}[c]{@{}l@{}}Medical\\      Datasets\end{tabular} & Modality                     & Target                                                                            & \begin{tabular}[c]{@{}c@{}}Testset\\      Num\end{tabular} \\ \midrule
  EndoVis18           \cite{allan20202018,gonzalez2020isinet}        & \multirow{2}{*}{Endoscopy}   & Instrument                                                                             & 1200                                                       \\
  LDPolypVideo     \cite{ma2021ldpolypvideo}                      &                              & Polyp                                                                             & 1040                                                       \\ \midrule
  ISIC16           \cite{gutman2016skin}                          & \multirow{3}{*}{Photography} & \begin{tabular}[c]{@{}c@{}}Skin\\      Lesions\end{tabular}                       & 379                                                        \\
  HAM10000         \cite{tschandl2018ham10000}                    &                              & \begin{tabular}[c]{@{}c@{}}Skin\\      Lesions\end{tabular}                       & 2000                                                       \\ \midrule
  TN3K             \cite{gong2023thyroid}                         & \multirow{2}{*}{Ultrasound}  & \multicolumn{1}{c}{\begin{tabular}[c]{@{}c@{}}Thyroid\\      Nodule\end{tabular}} & 614                                                        \\
  BUID             \cite{al-dhabyaniDatasetBreastUltrasound2020}  &                              & \multicolumn{1}{c}{\begin{tabular}[c]{@{}c@{}}Breast\\      Cancer\end{tabular}}  & 320                                                        \\ \midrule
  TBX11K           \cite{liu2020tbx11k}                           & \multirow{2}{*}{DR}          & Tuberculosis                                                                      & 1000                                                       \\
  RSNA  Pneumonia           \cite{shih2019augmenting}                      &                              & Pneumonia                                                                         & 1000                                                       \\ \midrule
  Luna16           \cite{setio2017validation}                     & \multirow{2}{*}{CT}          & \multicolumn{1}{c}{\begin{tabular}[c]{@{}c@{}}Lung\\      Nodule\end{tabular}}    & 125                                                        \\
  DeepLesion       \cite{yan2018deeplesion}                       &                              & Lesion                                                                            & 660                                                        \\ \midrule
  ADNI             \cite{muellerWaysEarlyDiagnosis2005}           & \multirow{2}{*}{MR}          & \multicolumn{1}{c}{Hippocampus}                                                   & 1700                                                       \\
  LGG              \cite{bakas2017advancing}                      &                              & Gliomas                                                                           & 680                                                        \\ \bottomrule
  \end{tabular}}
\end{table}

\begin{table}[htbp]
\caption{The training hyperparameters of our method.}
\label{table:param}
\centering
\setlength{\tabcolsep}{15mm}{
\begin{tabular}{@{}lc@{}}
\toprule
\multicolumn{2}{l}{\textbf{Hyperparameters}} \\ \midrule
Training Steps       & 70,000       \\
Warmup Steps         & 1,000        \\
Optimizer            & AdamW        \\
Learning Rate        & 2e-5         \\
Learning Rate Decay  & Cosine       \\
Adam $\beta$         & (0.9, 0.98)  \\
Weight Decay         & 0.05         \\
Batch Size           & 12           \\ \bottomrule
\end{tabular}
}
\end{table}

\begin{table*}[htbp]
  \caption{Evaluation results of visual grounding on RefCOCO, RefCOCO+ and RefCOCOg datasets. 
  Acc@0.5 is applied to evaluate the performance of two types of visual grounding methods, i.e., specialist and generalist model.
  }
  \label{tab:vg-results}
  \centering
  \setlength{\tabcolsep}{1mm}{
  \begin{tabular}{@{}cccccccccccc@{}}
  \toprule
&  & &&\multicolumn{3}{c}{RefCOCO}                                                                                           & \multicolumn{3}{c}{RefCOCO+}                                                                                          & \multicolumn{2}{c}{RefCOCOg}                                                  \\ \cmidrule(l){5-12} 
  \multirow{-2}{*}{Models} & \multirow{-2}{*}{Venue} &\multirow{-2}{*}{Visual Encoder}& \multirow{-2}{*}{Language Model}              & val                                   & testA                                 & testB                                 & val                                   & testA                                 & testB                                 & val-u                                 & test-u                                \\ \midrule
  \textbf{Specialist:}\\
  CM-A-E
\cite{liuImprovingReferringExpression2019} &CVPR19&ResNet-101& LSTM & 87.47  & 88.12   & 86.32                                 & 73.74                                 & 77.58                                 & 68.85                                 & 80.23                                 & 80.37  \\     NMTREE\cite{liuLearningAssembleNeural2019} &ICCV19&ResNet-101& Bi-LSTM           & 85.65                                 & 85.63                                 & 85.08                                 & 72.84                                 & 75.74                                 & 67.62                                 & 78.57                                 & 78.21                                                         \\
  DGA \cite{yangDynamicGraphAttention2019}  &ICCV19& ResNet-101    & Bi-LSTM                                      & 86.34                                 & 86.64                                 & 84.79                                 & 73.56                                 & 78.31                                 & 68.15                                 & 80.21                                 & 80.26                                 \\

  MCN \cite{luo2020multi}&CVPR20&DarkNet-53&Bi-GRU&80.08&82.29&74.98&67.16&72.86&57.31&66.46&66.01\\
  
  ReSC-Large                              \cite{yang2020improving}       &ECCV20&DarkNet-53& BERT-base     & 77.63                                 & 80.45                                 & 72.30                                 & 63.59                                 & 68.36                                 & 56.81                                 & 67.30                                 & 67.20                                 \\
  TransVG                                 \cite{Deng_2021_ICCV}  &ICCV21 &ResNet-101& BERT-base & 81.02                                 & 82.72                                 & 78.35                                 & 64.82                                 & 70.70                                 & 56.94                                 & 68.67                                 & 67.73                                 \\                         
  MDETR \cite{kamath2021mdetr}&ICCV21&EfficientNet-B3&RoBERTa-base                                       & 86.75                                 & 89.58                                 & 81.41                                 & 79.52                                 & 84.09                                 & 70.62                                 & 81.64                                 & 80.89                                 \\
  SeqTR \cite{Zhu_2022} & ECCV22 &DarkNet-53&Bi-GRU& 83.72                                 & 86.51                                 & 81.24                                 & 71.45                                 & 76.26                                 & 64.88                                 & 74.86                                 & 74.21                                 \\
  VGTR \cite{du2022visual} &ICME22&EfficientNet-B3&RoBERTa-base                                     & 79.30                                 & 82.16                                 & 74.38                                 & 64.40                                 & 70.85                                 & 55.84                                 & 66.83                                 & 67.28                                 \\
  \midrule
  \textbf{Generalist:}\\
 OFA \cite{wang2022ofa} &   ICML22 &ResNet-152&BART-Large& 92.04 & 94.03 & 88.44& {\color[HTML]{000000} 87.86} & 91.70 & 80.71  & 88.07 & 88.78 \\ 
mPLUG-2 \cite{xuMPLUG2ModularizedMultimodal2023}   &  ICML23  &ViT-L14& BERT-Large& 90.33  & 92.80 & 86.05   & - & -  & -& 84.70 & 85.14           \\
Kosmos-2  \cite{peng2023kosmos}    &ICLR24&ViT-L14& Magneto-1.3B&- & -  & -  & -  &- & -  &61.65  & 86.96 \\
Ferret  \cite{you2024ferret} & ICLR24 &ViT-L14&Vicuna-7B  & 87.49 & 91.35  & 82.45  & 80.78  &87.38 & 73.14  &83.93  & 84.76 \\
VistaLLM \cite{pramanick2023jack}      &CVPR24&EVA ViT&Vicuna-7B&88.10 & 91.50  & 83.00  & 82.90  &89.80 & 74.80  &83.60  & 84.40 \\
  RegionGPT \cite{guo2024regiongpt}      &CVPR24&ViT-L14&Vicuna-7B&- & -  & -  & -  &- & -  &60.57  & 86.96 \\
  Shikra \cite{chen2023shikra}&  arxiv 23.6    &ViT-L14&Vicuna-13B                              & 87.83                                 & 91.11                                 & 81.81                                 & 82.89                                 & 87.79                                 & 74.41                                 & 82.64                                 & 83.16                                 \\
  Qwen-VL \cite{bai2023qwen}   &arxiv 23.8&ViT-G&QWen-7B&88.55 & 92.27  & 84.51  & 82.82  &88.59 & 76.79  &85.96  & 86.32 \\
  COMM                                 \cite{jiang2023clip} &  arxiv 23.10 &CLIP+DINOv2&Vicuna-7B& 91.73   & 94.06 & 88.85& 87.21&91.74  & 81.39   & 87.32&88.33\\
  CogVLM-17B \cite{wang2023cogvlm}    &arxiv 23.11&EVA2-CLIP-E&Vicuna-7B&92.76 &  94.75 &  88.99 &  88.68 & 92.91&  83.39  & 89.75 & 90.79 \\
  MiniGPT-v2 \cite{chen2023minigpt} &arxiv 23.10&EVA ViT&LLaMA2-7B&88.06 & 91.29  & 84.30  & 79.58  &85.52 & 73.32  &84.19  & 84.31 \\
  NExT-Chat \cite{zhang2023next}   &arxiv 23.11&ViT-L14&Vicuna-7B&85.5 & 90.0  & 77.9  & 77.2  &84.5 & 68.0  &80.1  & 79.8 \\
  SPHINX-2K \cite{lin2023sphinx}    &arxiv 23.11& ViT-L14&LLaMA2&91.10 & 92.88  & 87.07  & 85.51  &90.62 & 80.45  &88.07  & 88.65 \\
  
  LLaVA-G \cite{zhang2023llava} &arxiv 23.12 &Swin Tiny&Vicuna-7B&89.16 & -  & -  & 81.68  & & -  & 84.82  & -  \\
  Ferret-v2 \cite{zhang2024ferret}&arxiv 24.4&ViT-L14&Vicuna-7B&  92.79 &  94.68  & 88.69  & 87.35  & 92.75 & 79.3  & 89.42 & 89.27  \\
  
  \textbf{Ours}&&ViT-L14 &Vicuna-7B & 92.99& 95.90& 90.39& 90.96&  94.78 & 86.93&  90.05&  89.51\\ 
   
  \bottomrule
  \end{tabular}}
\end{table*}

\section{Experimental Details}

\subsection{Datasets}
RefCOCO \cite{yuModelingContextReferring2016}, RefCOCO+ \cite{yuModelingContextReferring2016}, and RefCOCOg \cite{maoGenerationComprehensionUnambiguous2016} are three visual grounding datasets that utilize images sourced from MSCOCO \cite{linMicrosoftCOCOCommon2014}. In line with previous approaches, we adopt the train / validation / testA / testB split for both RefCOCO and RefCOCO+ datasets, where testA and testB sets contain only people and only non-people respectively. The split of RefCOCOg-umd \cite{nagarajaModelingContextObjects2016} on RefCOCOg refers to the splits as the val-u, and test-u. Accuracy@0.5 (Acc@0.5) is used to measure the performance of the visual grounding task, which is right if the IoU between the grounding-truth box and the predicted bounding box is larger than 0.5.

To evaluate the model's generalization capabilities in the medical field, we test its performance on public datasets for disease identification across 6 modalities, namely endoscopy, photography, ultrasound, DR, CT, and MRI, to assess its robustness and adaptability.

The Object-CXR \cite{objectcxr} dataset is designed for the automatic detection of foreign objects in chest X-rays. It consists of 5,000 frontal chest X-ray images with foreign objects and 5,000 images without foreign objects. These DR images were captured and collected from approximately 300 township hospitals in China. The ChestXray14 dataset \cite{wang2017chestx} contains 112,120 chest X-ray images with labels for 14 common diseases. Among these, 984 images feature eight key findings with hand-labelled bounding boxes.
The RSNA Pneumonia dataset \cite{shih2019augmenting} is a binary classification chest X-ray dataset consisting of 26,683 images. Each radiograph is categorized as either pneumonia or normal.
The TBX11K dataset \cite{liu2020tbx11k} is a large collection comprising 11,000 chest X-ray images, each with corresponding bounding box annotations for tuberculosis areas.

Moreover, we conduct extensive experiments to evaluate the generalization capability of our model on various medical multi-modality datasets, as shown in Table.\ref{table:datasets}. 
EndoVis18  \cite{allan20202018}  is a publicly available dataset for endoscopy image analysis. We follow ISINet’s annotation and data set division of surgical instrument categories\cite{gonzalez2020isinet}.
LDPolypVideo \cite{ma2021ldpolypvideo}  consists of 44 colonoscopy videos for polyp detection, with a total of 18,142 frames, and a resolution of 512$\times$512 pixels. 
ISIC16 \cite{gutman2016skin} is a collection of dermoscopic images of skin lesions, annotated by dermatologists and skin cancer experts. It consists of 1,267 dermoscopic images of skin lesions, including melanomas and benign lesions, with a resolution of 1024$\times$768 pixels.
HAM10000 \cite{tschandl2018ham10000} dataset is a large, publicly available dataset for skin lesion analysis, specifically designed for melanoma detection and skin disease diagnosis. 
TN3K \cite{gong2023thyroid} dataset consists of 2D ultrasound images of thyroid nodules with a resolution of 512$\times$512 for thyroid nodules detection.
BUID \cite{al-dhabyaniDatasetBreastUltrasound2020} dataset consists of 780 images with an average image size of 500$\times$500 pixels from 600 female patients for breast cancer detection.
Luna16    \cite{setio2017validation} dataset is a publicly available dataset for lung nodule analysis, specifically designed for lung nodule detection in CT scans.
DeepLesion  \cite{yan2018deeplesion} dataset is a large-scale, publicly available dataset for lesion detection and segmentation in CT, with a resolution of 512x512 pixels. 
ADNI \cite{muellerWaysEarlyDiagnosis2005}  is a large, publicly available dataset for Alzheimer's disease research from magnetic resonance imaging (MRI) scans, specifically designed for the development and evaluation of algorithms for early detection and diagnosis of Alzheimer's disease. 
LGG  \cite{bakas2017advancing} (Low-Grade Glioma) dataset is a publicly available dataset for brain tumor segmentation, specifically for detecting low-grade gliomas from MRI scans. 

We strictly follow the official train-test split for EndoVis18, LDPolypVideo, ISIC16, TN3k, Luna16 and ADNI. Due to the absence of an official training/validation/test ratio or a released test set for HAM10000, BUID, TBX11K, RSNA Pneumonia, DeepLesion and LGG, we randomly split each dataset into training/validation/test sets by 7:1:2 for the visual grounding task.

\subsection{Implementation Details}

For our language-guided image tokenizer, we leverage the strengths of both BERT\cite{devlinBERTPretrainingDeep2019} and ViT as our text encoder and visual encoder, respectively. 
We employ ViT-L14 as our visual encoder, which consists of 14 transformer encoder layers and an FFN intermediate size of 4,096. The input image size is set to $224\times 224$, with a patch size of 16×16. The hidden dimensions of the ViT-L14 are 1,024, with 16 attention heads.
Meanwhile, we utilize Vicuna-7B, a large language model fine-tuned with instructions, as our text encoder. The Vicuna-7B model boasts 12 transformer layers, with 768 hidden dimensions, 12 attention heads, and an FFN intermediate size of 3,072. The vocabulary size is 30,522, and the maximum input sequence length is 512.
To align the text encoder and visual encoder, we employ a Q-former with 12 transformer layers. This Q-former has 768 hidden dimensions, 12 attention heads, and query, key, and value dimensions of 256 each.

In terms of the training progress, the hyperparameters are presented in Table.\ref{table:param}. 
We utilize the AdamW optimizer, which is configured with a cosine annealing schedule as the learning policy. The initial learning rate is set to $2\times10^{-5}$, and the AdamW optimizer is employed with hyperparameters $\beta= (0.9, 0.98)$. Additionally, we set the weight decay to 0.05 and the dropout rate to 0.1. During the first 1,000 warm-up steps, the learning rate increases to $2\times10^{-5}$, and subsequently decays to $10^{-7}$. Unless otherwise specified, our training protocol consists of 70,000 steps, executed on $4\times 8$ NVIDIA V100 GPUs, which takes approximately two days to complete.

For the annotation, We normalize all coordinates to a uniform range of 0 to 1000, ensuring that all images have a consistent coordinate system. 
For the polygon representation, we select the point closest to the origin as the starting point and employ a 25-point labelling scheme to describe the polygon sequence in a clockwise direction. To demarcate the beginning and end of the sequence, we utilize <BOS> and <EOS> tags, respectively.
For the sampling rule for polygons, we employ isometric sampling, wherein we initially calculate the perimeter of the polygon and subsequently divide it into 25 equal segments to sample the polygon.

\section{Additional Experiment Results}

\begin{table}[htbp]
  \caption{Evaluation results of referring object classification on LVIS and COCO 2017 val set. 
  ACC is utilized to validate the performance of referring object classification.}
  \label{table:objectCls}
  \centering
  \begin{tabular}{@{}lccccc@{}}
      \toprule
      \multicolumn{1}{c}{}                            &   \multicolumn{3}{c}{LVIS}                                                                              
      & \multicolumn{1}{c}{COCO 17}     
      \\ \cmidrule(l){2-5} 
      \multicolumn{1}{c}{\multirow{-2}{*}{Methods}}                      
      & keypoints                                   & bbox                                  & polygon &bbox                                   \\ \midrule
LLaVA \cite{wu2022grit} &50.10 & 50.30 &- &40.04\\
Shikra\cite{chen2023shikra} &57.82&67.71&-&53.91\\
Ferret \cite{you2024ferret} &67.94& 79.42 & 69.77\\
      Kosmos-2     \cite{peng2023kosmos}     &  -                       &60.25      &-    \\
      GPT4RoI           \cite{zhang2023gpt4roi}               &  -    &61.76         & -             \\
      $\textbf{Ours}$  & 58.24 & 66.42& 67.00&80.58  \\    
      \bottomrule
  \end{tabular}
                                 
\end{table}

\subsection{Referring Object Classification Task}
The performance of object classification was evaluated on LVIS and COCO 2017 datasets using classification accuracy. As shown in Table.\ref{table:objectCls}, the referring object classification task of our method achieved excellent results using different visual prompts, i.e., keypoints, bbox, and polygon, narrowly surpassed only by the Ferret method on the LVIS dataset.
For the COCO 17 dataset, our method yield better performance than LLaVA and Shikra.
The comparison results for other methods on the LVIS and COCO 17 datasets in Table.\ref{table:objectCls} are sourced from the Ferret \cite{you2024ferret}and PVIT \cite{chen2023position} papers, respectively.

\subsection{VGCoco}

In response to the growing need for high-quality referring expression data that captures diverse forms of visual location, we introduce VGCoco, a comprehensive dataset designed to meet this demand. Comprising approximately 240,000 images, VGCoco features a range of grounding annotations, from region-level to pixel-level, accompanied by corresponding referring expressions.

We build upon existing open-source datasets, including COCO-Pose \cite{lin2015microsoft}, AP-10K \cite{yu2021ap}, and a subset of COCO 2017 \cite{linMicrosoftCOCOCommon2014}. Notably, we leverage the cycle training strategy to generate the reference expressions required by COCO-Pose using our model.
COCO-Pose provides a wealth of the keypoints information on the human body with the skeleton for pose estimation, as well as bounding boxes and segmentated polygons, but lacks referring expressions. Therefore, we employ our model to generate the detailed referring expressions of persons in COCO-Pose. 
Besides, AP-10K is the animal version of COCO-Pose, which consists of keypoints of animal skeletons. With the aid of SAM\cite{kirillovSegmentAnything2023}, we get the masks of the target animal and transfer them to polygon by uniform sampling. Then, our model is exploited to produce referring expressions for the target animal. Furthermore, we randomly selected a part of non-people and non-animal targets in COCO 2017 to build the VGCoco dataset. In addition to using our model to generate key points, we use the skeletonization method to obtain the key points of these targets.
\begin{table*}[htbp]
\caption{Evaluation results of \textbf{referring bbox detection} task on 12 typical medical datasets of 6 modalities. 20-shot fine-tuning experiments were performed for non-radiology (Endoscopy, Photography and Ultrasound) and radiology datasets (DR, CT and MRI). Acc@0.5 is applied to evaluate methods.}
  \label{tab:medicalbenchmark}    

  \centering
  \setlength{\tabcolsep}{0.8mm}{
  \begin{tabular}{@{}lcccccccccccc@{}}
  \toprule
  \multicolumn{1}{c}{\multirow{2}{*}{Datasets}}   & \multicolumn{2}{c}{Endoscopy}  & \multicolumn{2}{c}{Photography}       & \multicolumn{2}{c}{Ultrasound} & \multicolumn{2}{c}{DR}     &\multicolumn{2}{c}{CT}     & \multicolumn{2}{c}{MRI}                                       \\
    \cmidrule(l){2-13}
  \multicolumn{1}{c}{} &EndoVis18 
  & LDPolypVideo &ISIC16&HAM10000&TN3K&BUID& TBX11K&RSNA&Luna16 & DeepLesion & ADNI&LGG \\ \midrule
  VGTR \cite{du2022visual}   &3.87
  &7.30&64.12&63.20&12.70&31.46&1.99&4.67&0.00&0.36&2.46&3.67    \\ 
  OFA    \cite{wang2022ofa}  &7.32
&0.30&63.85&61.20&6.81&19.62&20.40&14.67&0.00&2.08&26.26&26.77  \\
  Ours   &12.53
  & 9.86 & 67.66 & 86.00 & 16.50 & 38.63&30.84&28.00 & 0.00 & 5.23 & 4.52 & 18.85   \\ 
   \bottomrule
  \end{tabular}
  }
\end{table*}

\subsection{Ablation Experiments}

We implement ablation experiments to validate the effect of cycle training on the referring region caption task. As shown in Table.\ref{table:Refcaption}, regardless of whether keypoints, bounding boxes, or polygons are used as visual prompts, cycle training consistently enhances captioning performance, with improvements of about 5\% and 15\% in METEOR and CIDEr scores, respectively. 

\begin{table}[htbp]
  \caption{Ablation experiment results of referring region caption on RefCOCOg. METEOR and CIDEr are utilized to validate the performance of region captioning.}
  \label{table:Refcaption}
  \centering
  \begin{tabular}{@{}lccccc@{}}
      \toprule
      \multicolumn{1}{c}{}                                                &   \multicolumn{2}{c}{With cycle training}                                                 & \multicolumn{2}{c}{Without cycle training}                                                                                               \\ \cmidrule(l){2-5} 
      \multicolumn{1}{c}{\multirow{-2}{*}{Visual Prompt}}  
      & METEOR                                   & CIDEr                                  & METEOR& CIDEr                                     \\ \midrule
       $\textbf{keyoints}$  & 26.5  & 164.6  & 19.4 & 149.3 \\
       $\textbf{bbox}$      & 26.3  & 167.2  & 19.1 & 152.7 \\ 
       $\textbf{polygon}$   & 26.6  & 165.5  & 19.8 & 151.2 \\
      \bottomrule
  \end{tabular}
                                 
\end{table}


\begin{table}[htbp]
\caption{Evaluation results of \textbf{referring bbox detection} task with 20-shot setting on seven labels from the EndoVis18 dataset. Acc@0.5 is applied to evaluate methods. Seven surgical instruments contain Bipolar Forceps (BF), Prograsp Forceps (PF), Large Needle Driver (LND), Monopolar Curved Scissors (MCS), Ultrasound Probe (UP), Suction Instrument (SI) and Clip Applier (CA).}
  \centering
  \setlength{\tabcolsep}{1.2mm}{
  \begin{tabular}{@{}lcccccccc@{}}
  \toprule
  Label & Mean &BF& PF & LND & MCS&UP &SI &CA\\ \midrule
  VGTR \cite{du2022visual}&3.87&12.29&0.00&0.00&14.78&0.00&0.00&0.00 \\ 
  OFA \cite{wang2022ofa}  &7.32&22.49&19.73&0.94&4.85&0.00&0.00&3.22 \\
  Ours  &12.53 & 16.88 & 4.17 & 3.33 & 13.55 & 0.00 & 0.00 & 16.67\\ 
   \bottomrule
  \end{tabular}
  
  }

  \label{tab:endo18}
\end{table}


\begin{figure}[htbp]
  \centering
  \begin{subfigure}[t]{.45\textwidth}
      \centering
      \includegraphics[width=\textwidth]{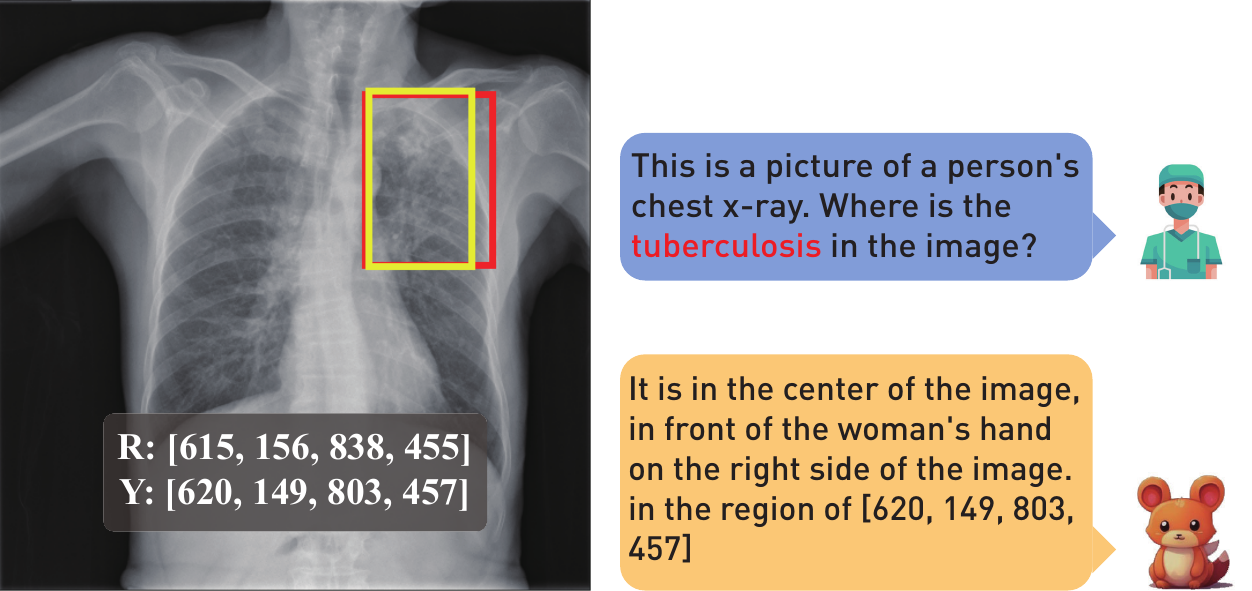}
      \caption{VQA in chest X-ray for the tuberculosis localization in the TBX11K dataset, without referring expressions.}
      \label{fig:tbx11k}
  \end{subfigure}
  \begin{subfigure}[t]{.45\textwidth}
      \centering
      \includegraphics[width=\textwidth]{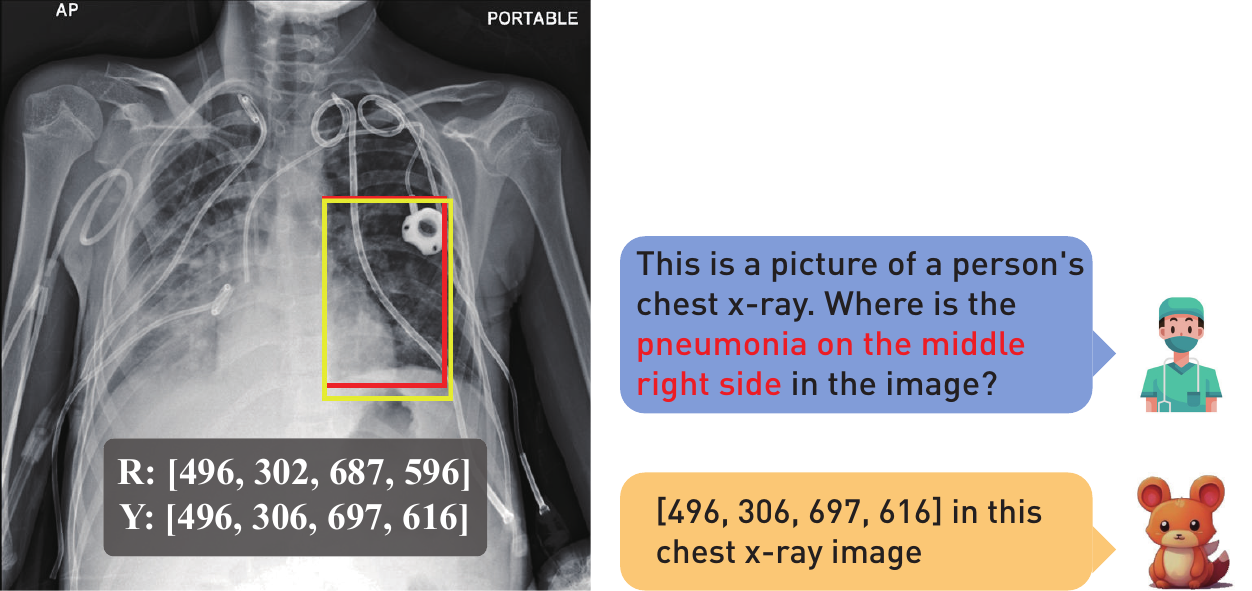}
      \caption{VQA in chest X-ray for the pneumonia localization in the RSNA Pneumonia dataset, with the orientation-related referring expression.}
      \label{fig:rsna}
  \end{subfigure}
  \caption{The 20-shot results of disease localization in chest X-ray images. The red box denotes the grounding truth, and the yellow box represents the prediction. (a) Tuberculosis detection in the TBX11K dataset. (b) Pneumonia detection in the RSNA dataset. }
  \label{fig:localization}
\end{figure}

\begin{figure*}[htbp]
  \centering
  \includegraphics[width=\textwidth]{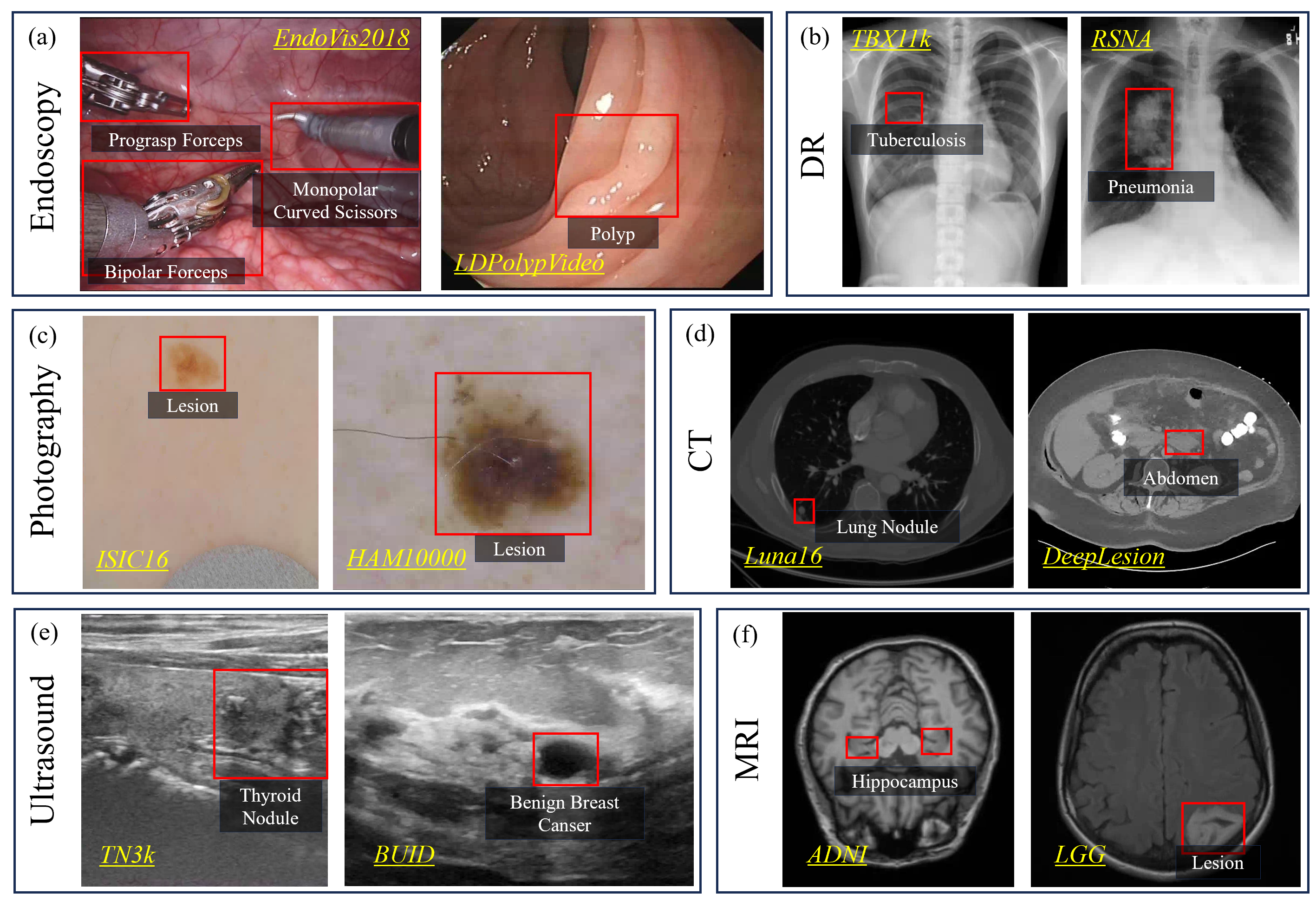}
  \caption{Typical medical datasets for referring bbox detection task, containing 6 modalities: (a) Endoscopy: EndoVis18, LDPolypVideo; (b) DR: TBX11k, RSNA Pneumonia; (c) Photography: ISIC16, HAM10000; (d) CT: Luna16, Deeplesion; (f) MRI: ADNI, LGG}
  \label{fig:medData}
\end{figure*}

\subsection{Generalization Performance in the Medical Domain}

To further examine the generalization and scalability of our model, we conduct preliminary experiments of the referring bbox detection task on 12 typical datasets across 6 modalities in the medical domain. The proposed model is fine-tuned with 20-shot labels for each disease of non-radiology and radiology datasets, respectively. We conduct comparative experiments with VGTR and OFA(Large) as representatives of specialist and generalist models, respectively, to evaluate their performance and versatility in the referring bbox detection task.

\subsubsection{Non-radiology Images} \ 

\textbf{Endoscopy}
We evaluated the generalization performance of instrument and disease localization on two typical endoscopy datasets, namely, EndoVis18 and LDPolyVideo.
As depicted in Table.~\ref{tab:medicalbenchmark}, 
ViLaM consistently outperforms other approaches. 
Table.~\ref{tab:endo18} further demonstrates that the proposed method achieves superior performance in multiple surgical instrument categories. However, there is still room for improvement in some categories, which may be attributed to the issue of data imbalance.

\textbf{Photography}
We evaluated the visual grounding performance of three methods on two datasets, ISIC16 and HAM10000, and found that all three methods achieved an accuracy of over 60\% on both datasets. This is likely due to the fact that skin disease images and their corresponding features share similar characteristics.

\textbf{Ultrasound}
We compared the visual grounding performance of three methods on two ultrasound datasets, TN3K and BUID, and our method achieved the best results. Specifically, our method achieved an accuracy of 16.50\% on the TN3K validation set and 38.63\% on the BUID breast cancer dataset.

\subsubsection{Radiology Images}\

\textbf{Chest X-ray}
Fig.\ref{fig:localization} illustrates the application of VQA in chest X-ray analysis for the localization of tuberculosis and pneumonia. This demonstrates that our generalist model effectively scales to the medical field, and it can adapt to medical disease localization tasks, with or without the use of referring expressions.
Quantitatively, our approach demonstrated a significant advantage over VGTR and OFA, with an improvement of over 10\% on the TBX11K and RSNA Pneumonia datasets.

\textbf{CT}
We compared the visual grounding performance of three methods on two CT datasets, Luna16 and DeepLesion, and found that all three methods achieved nearly 0\% accuracy in the 20-shot finetuning experiment on both datasets. This is likely due to the fact that the features of CT images and general images are quite different, and the lesions, such as lung nodules, are too small, as shown in Fig.\ref{fig:medData} (d).

\textbf{MRI}
Two MRI datasets, the ADNI dataset and the LGG dataset, verify the visual grounding performance of three methods. For Gliomas with larger contrast in the LGG dataset, we have better performance than VGTR. Our result of hippocampus detection is poor due to the low contrast of ADNI, as illustrated in Fig.\ref{fig:medData} (f).


\subsection{Qualitative Examples}

\begin{figure*}[htbp]
  \centering
  \begin{subfigure}[t]{0.33\textwidth}
  \centering
  \includegraphics[width=0.99\textwidth]{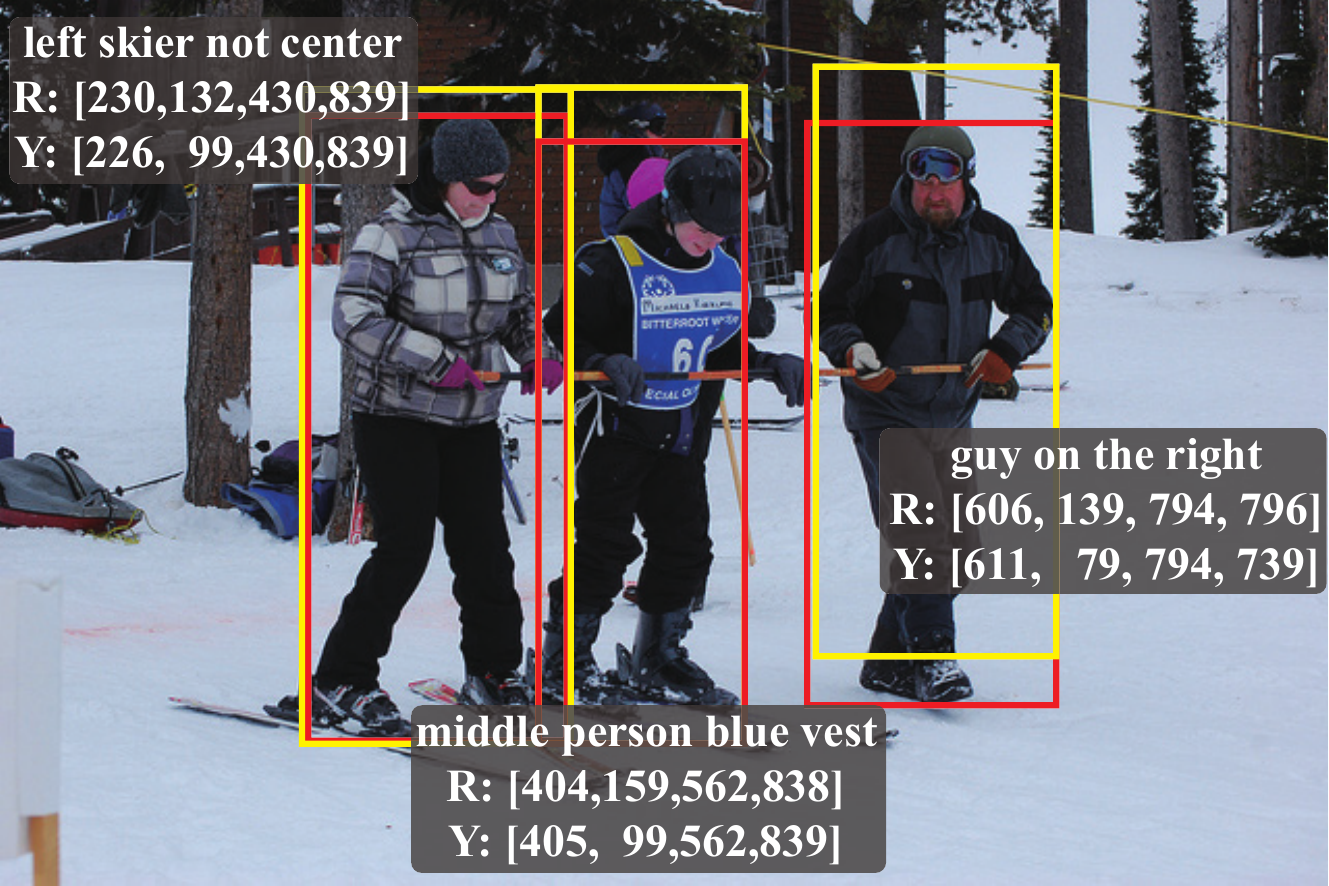}
  \end{subfigure}
  \begin{subfigure}[t]{0.33\textwidth}
  \centering
  \includegraphics[width=0.99\textwidth]{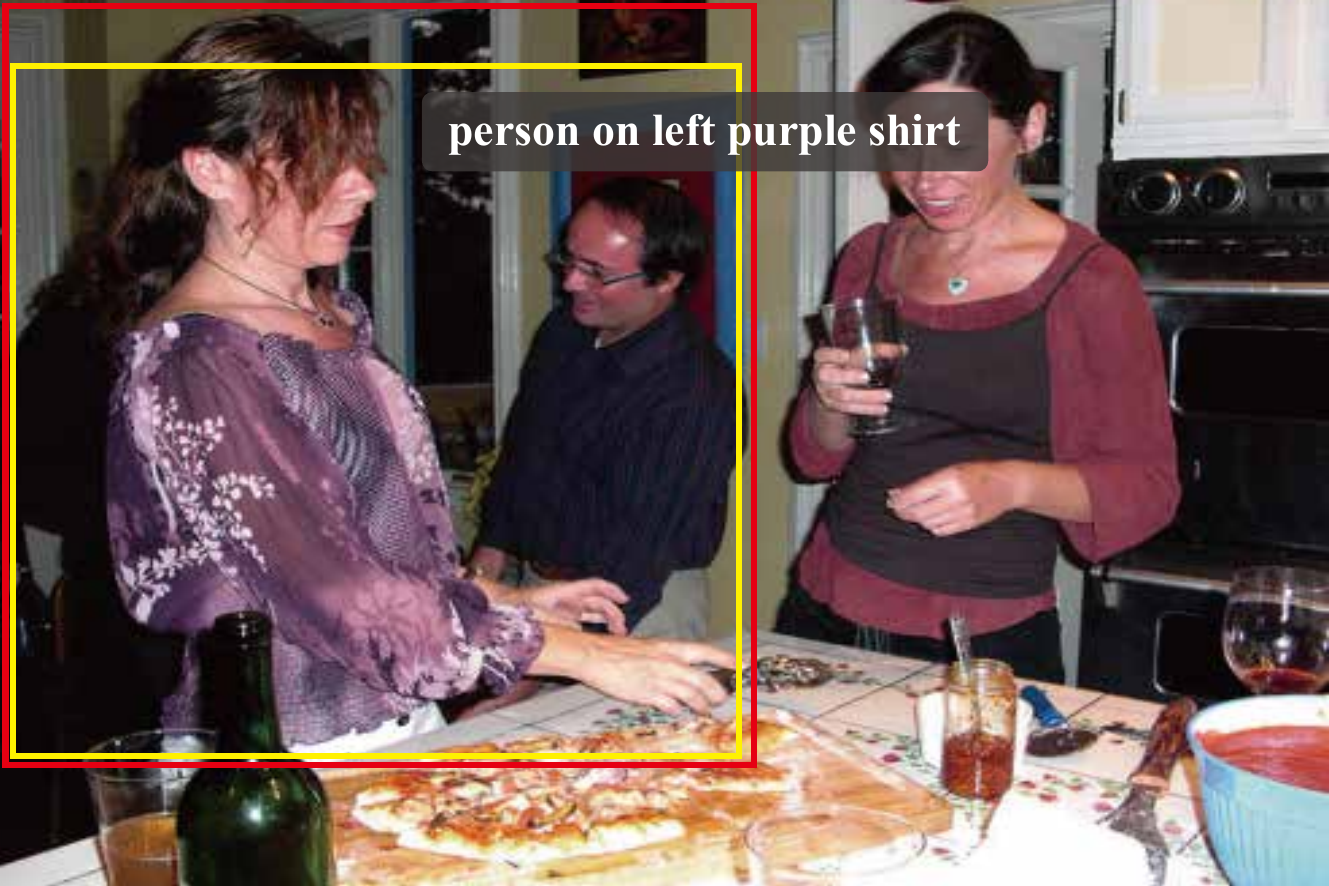}
  \end{subfigure}
  \begin{subfigure}[t]{0.33\textwidth}
  \centering
  \includegraphics[width=0.99\textwidth]{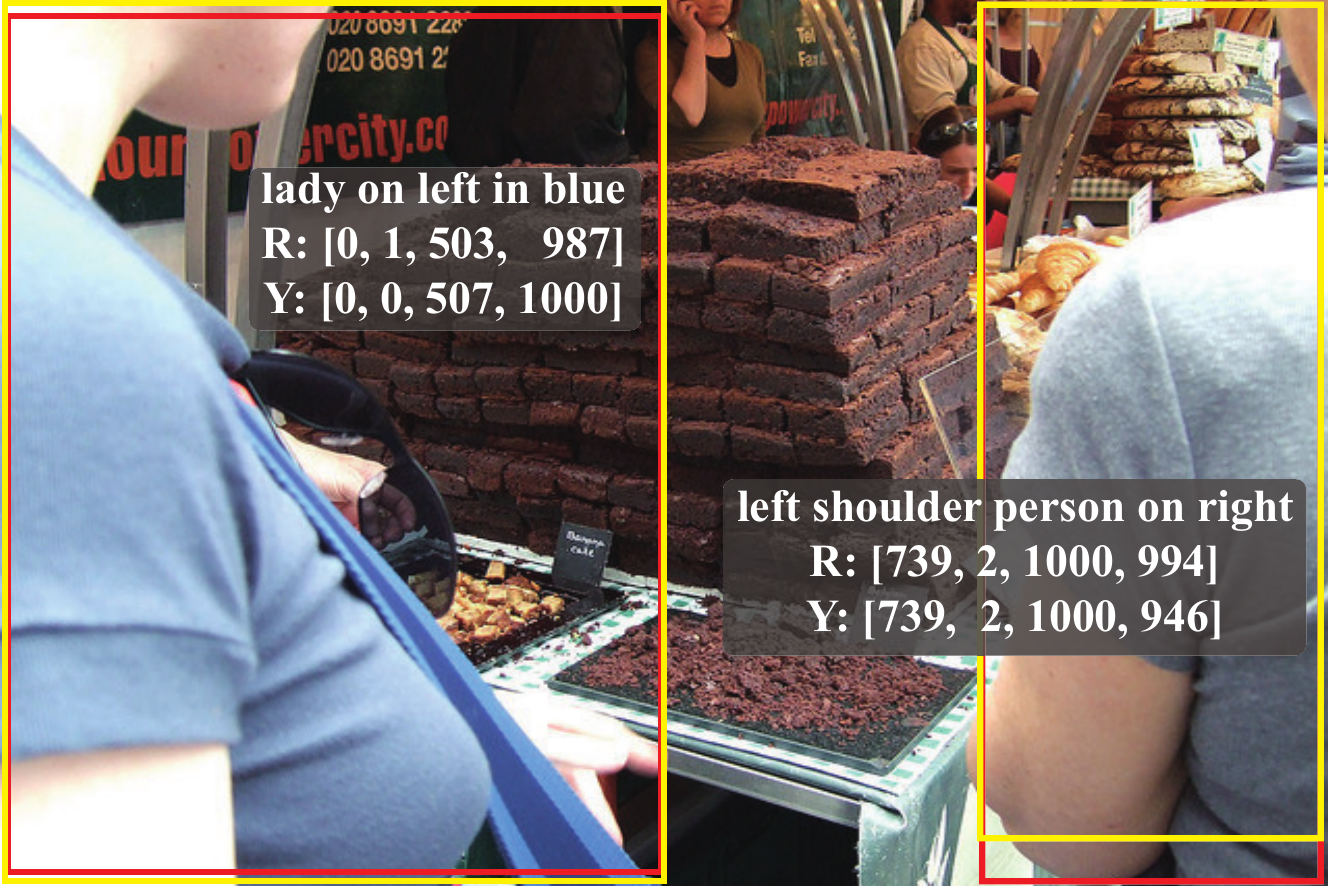}
  \end{subfigure}

  \begin{subfigure}[t]{0.33\textwidth}
  \centering
  \includegraphics[width=0.99\textwidth]{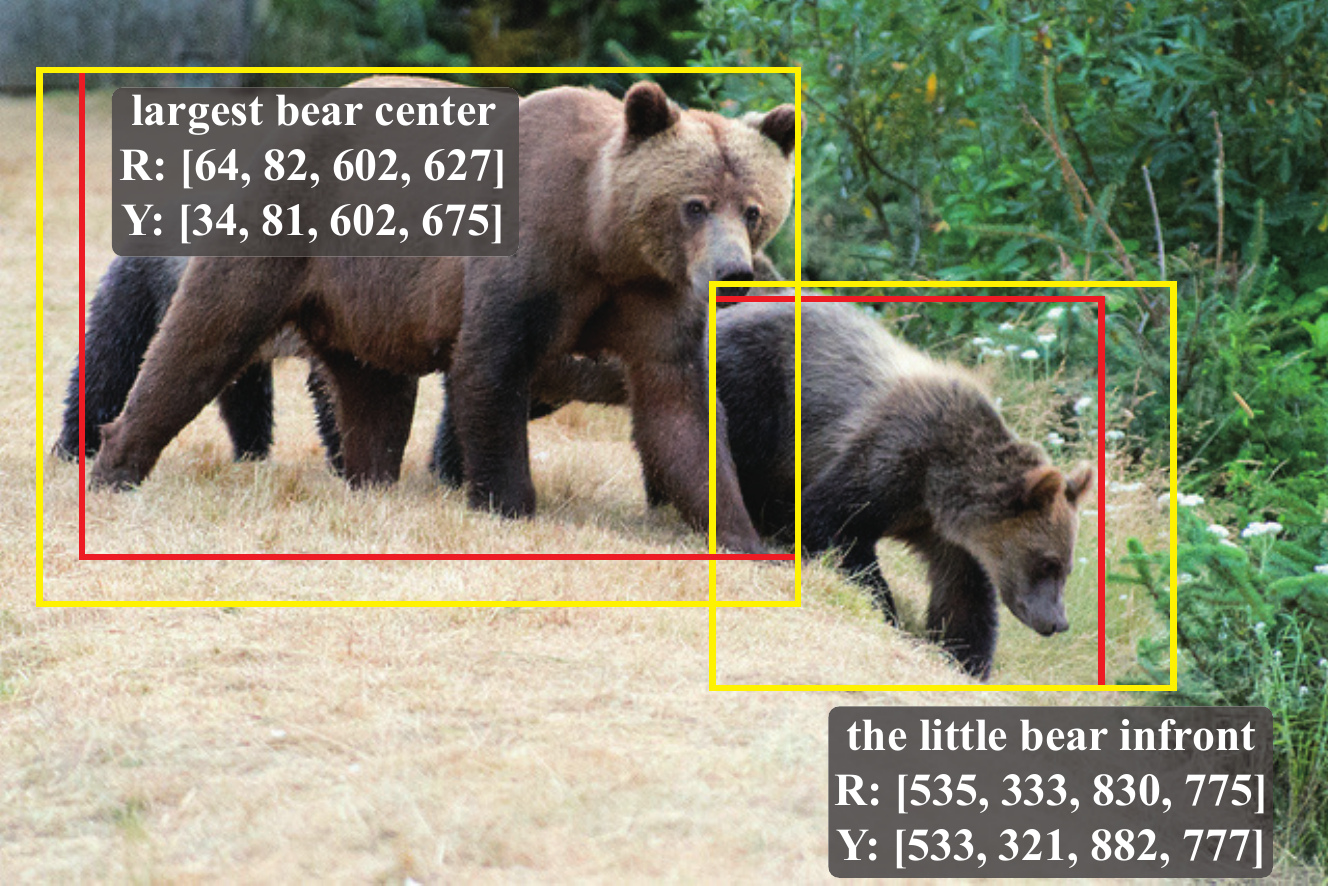}
  \end{subfigure}
  \begin{subfigure}[t]{0.33\textwidth}
  \centering
  \includegraphics[width=0.99\textwidth]{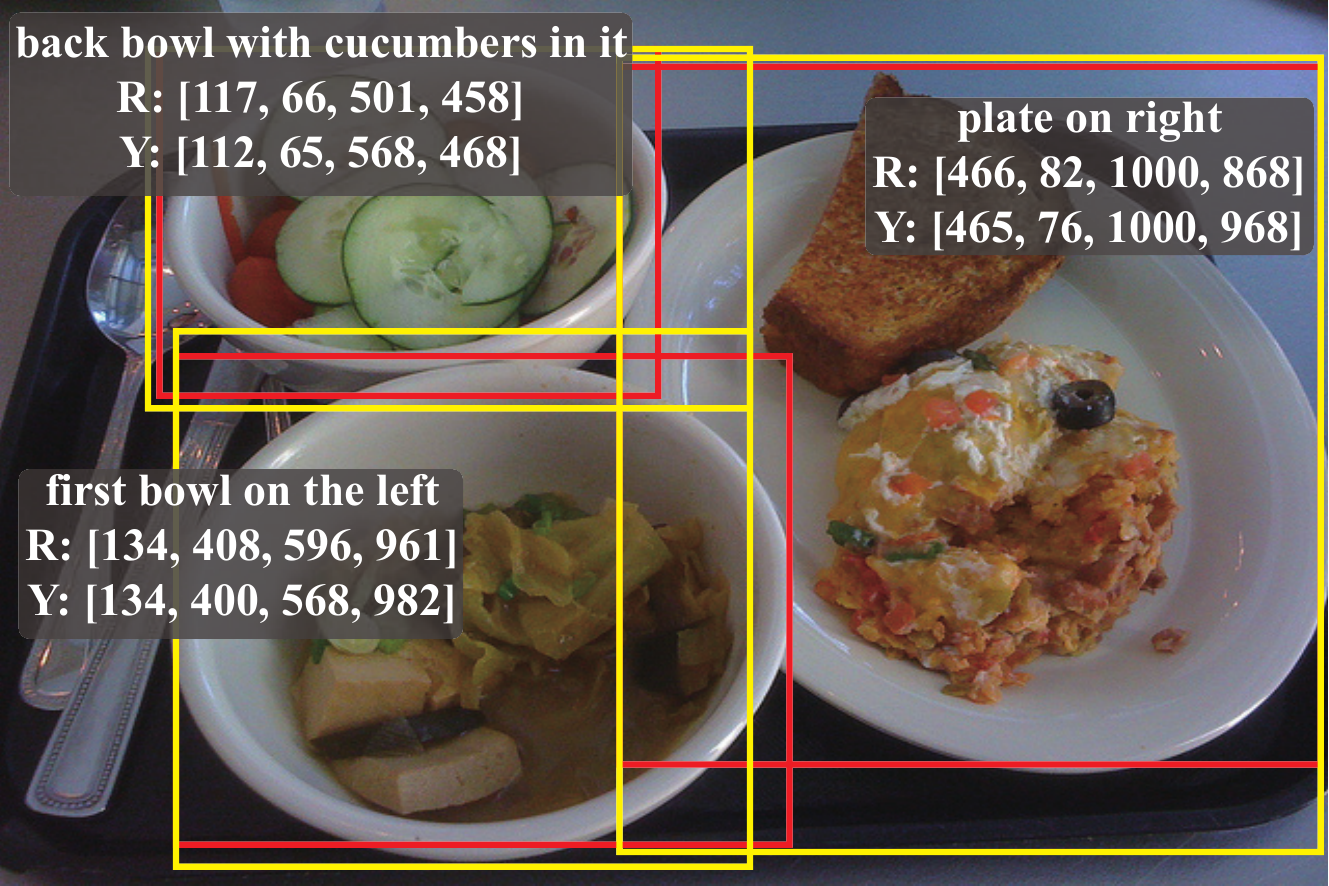}
  \end{subfigure}
  \begin{subfigure}[t]{0.33\textwidth}
  \centering
  \includegraphics[width=0.99\textwidth]{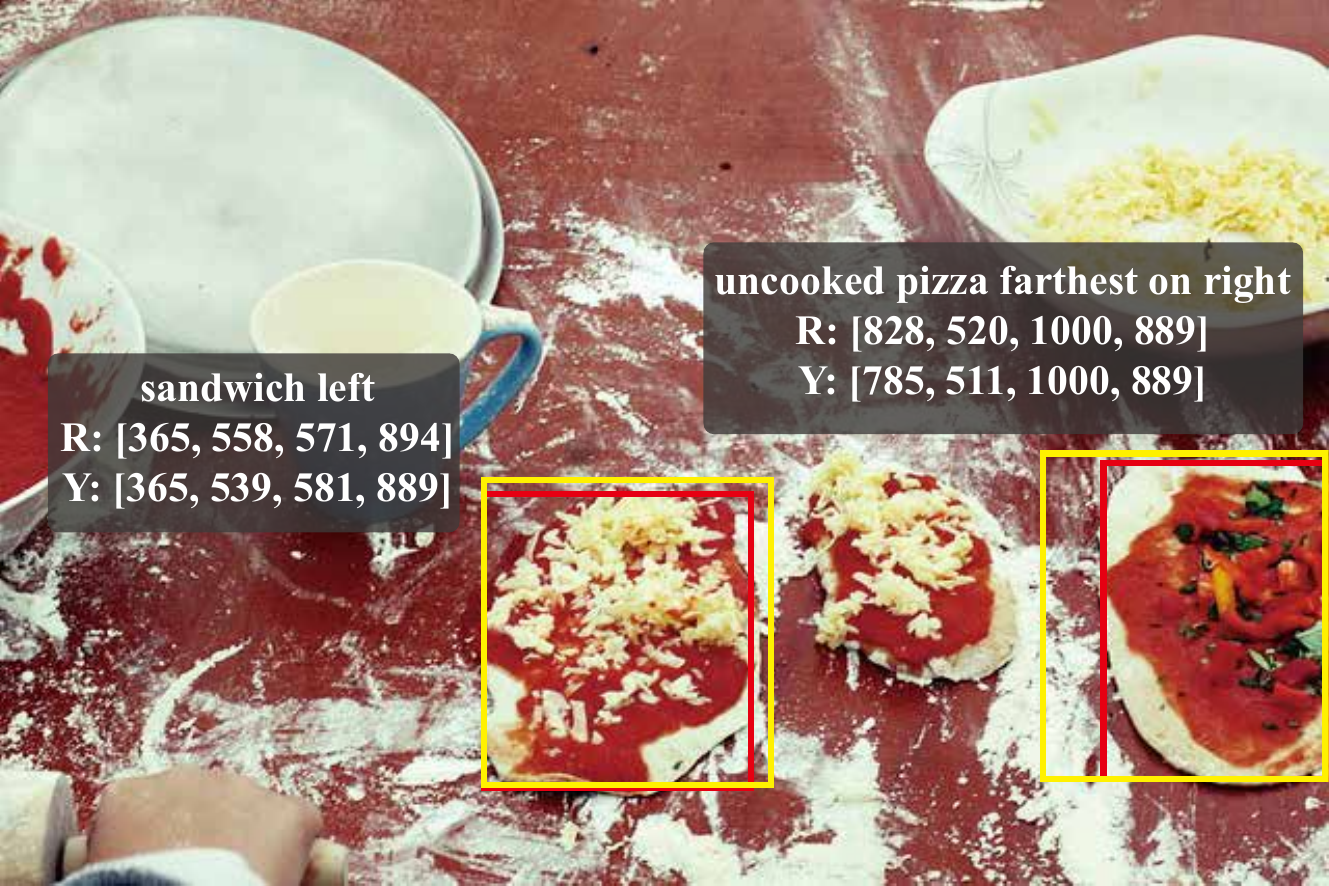}
  \end{subfigure}
  \caption{More Results of referring bbox detection in RefCOCO. The images of the first row are from testA split containing only people, while the second row images from testB consisting of only non-people. We display typical cases of referring expressions, especially with common indications of orientation, size, color, attachment and markings. The referring expressions of the object are presented in the text box with two coordinates, where R (red) denotes grounding truth and Y (yellow) symbolizes the prediction. The red and yellow bounding boxes are also depicted in the image, respectively. }
  \label{fig:bbox}
\end{figure*}

\begin{figure*}[htbp]
  \centering
  \begin{subfigure}[t]{0.33\textwidth}
  \centering
  \includegraphics[width=0.99\textwidth]{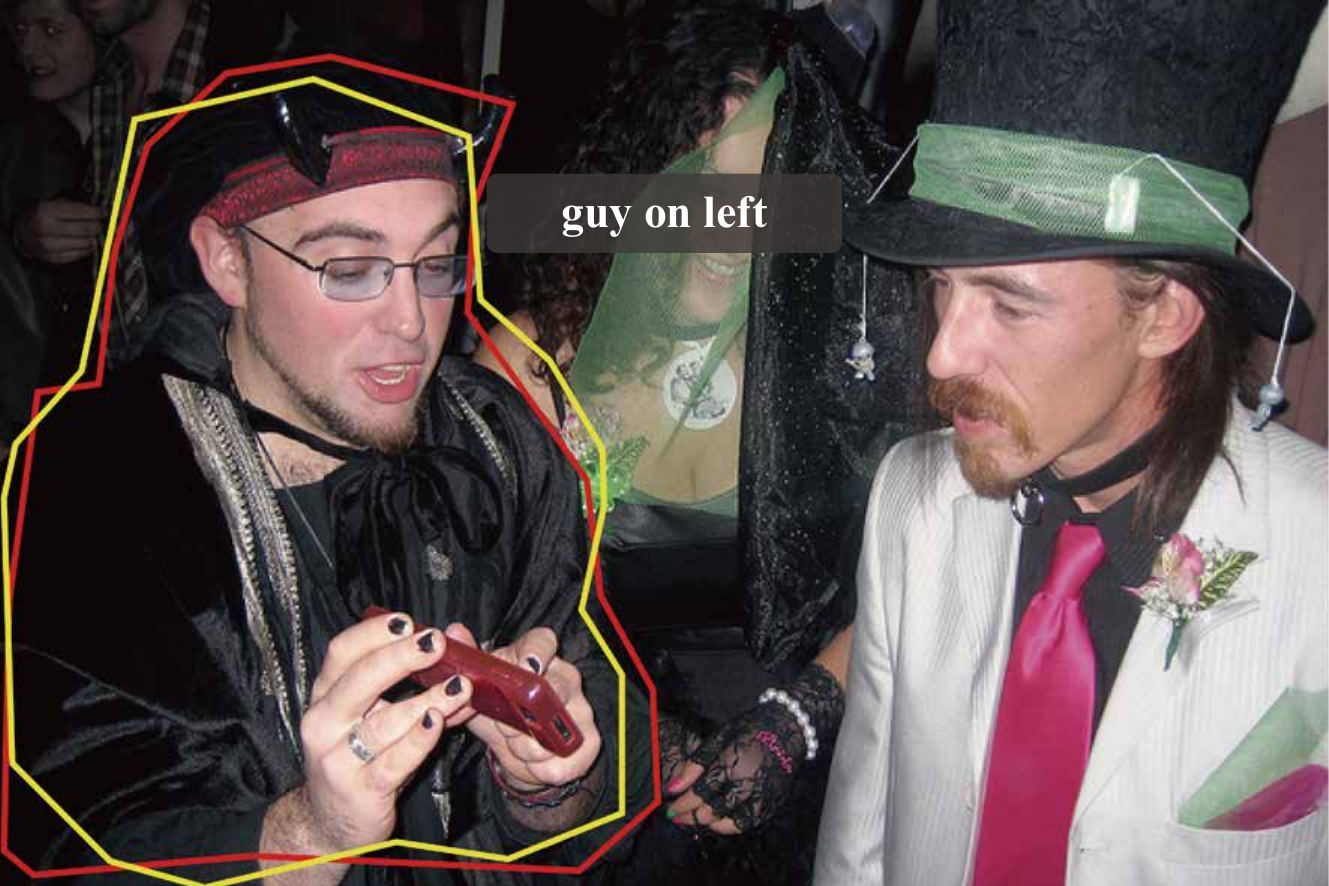}
  \end{subfigure}
  \begin{subfigure}[t]{0.33\textwidth}
  \centering
  \includegraphics[width=0.99\textwidth]{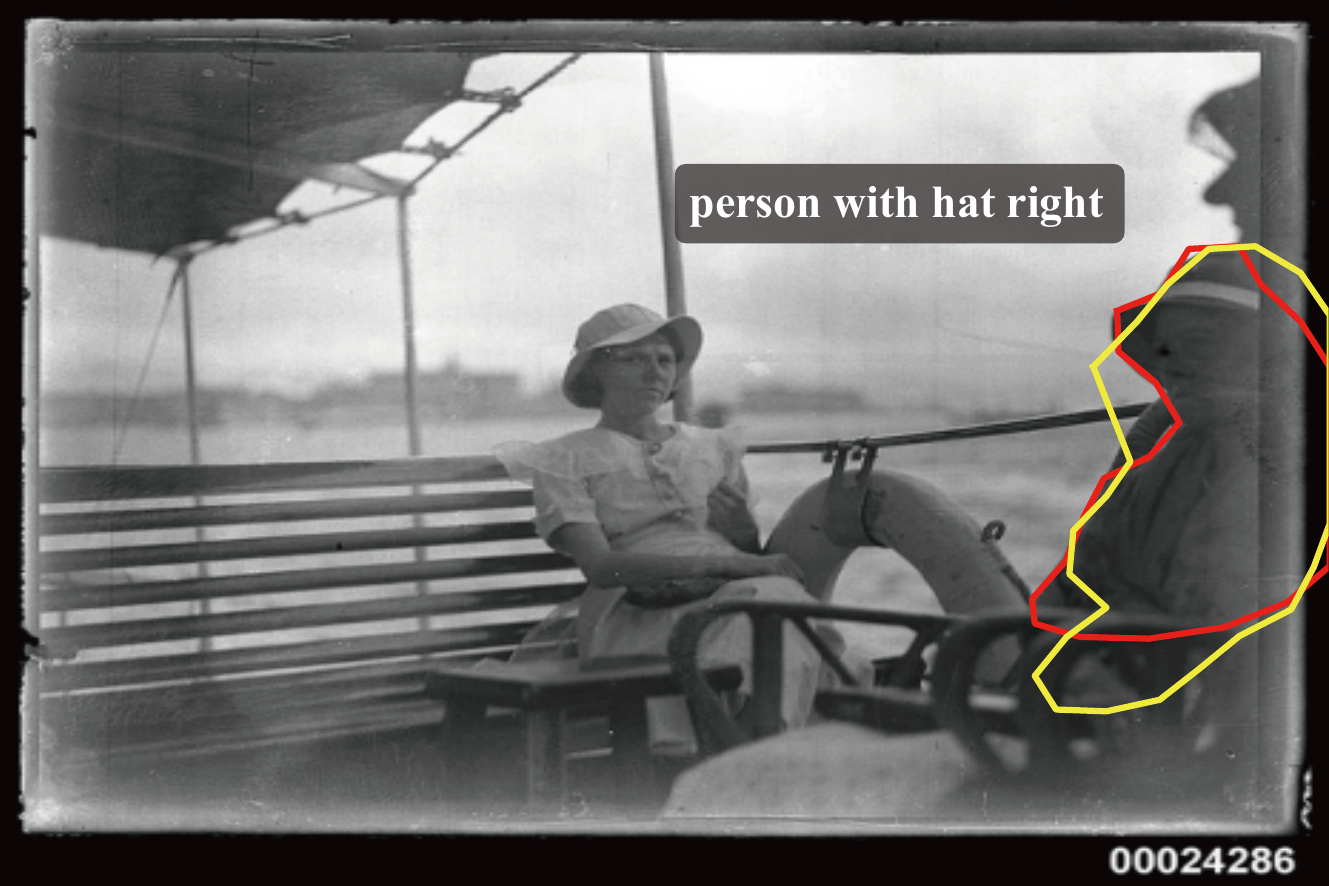}
  \end{subfigure}
  \begin{subfigure}[t]{0.33\textwidth}
  \centering
  \includegraphics[width=0.99\textwidth]{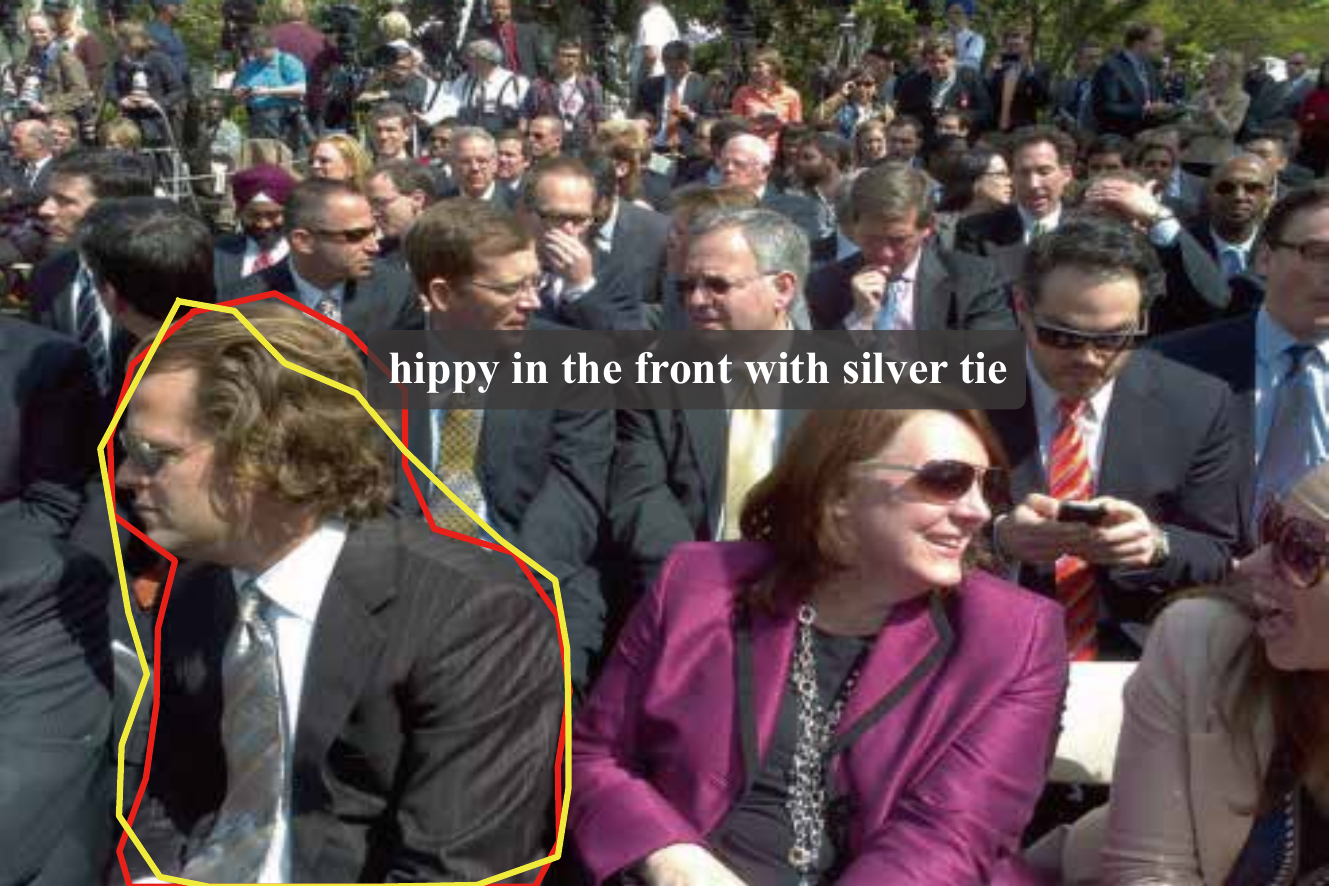}
  \end{subfigure}

  \begin{subfigure}[t]{0.33\textwidth}
  \centering
  \includegraphics[width=0.99\textwidth]{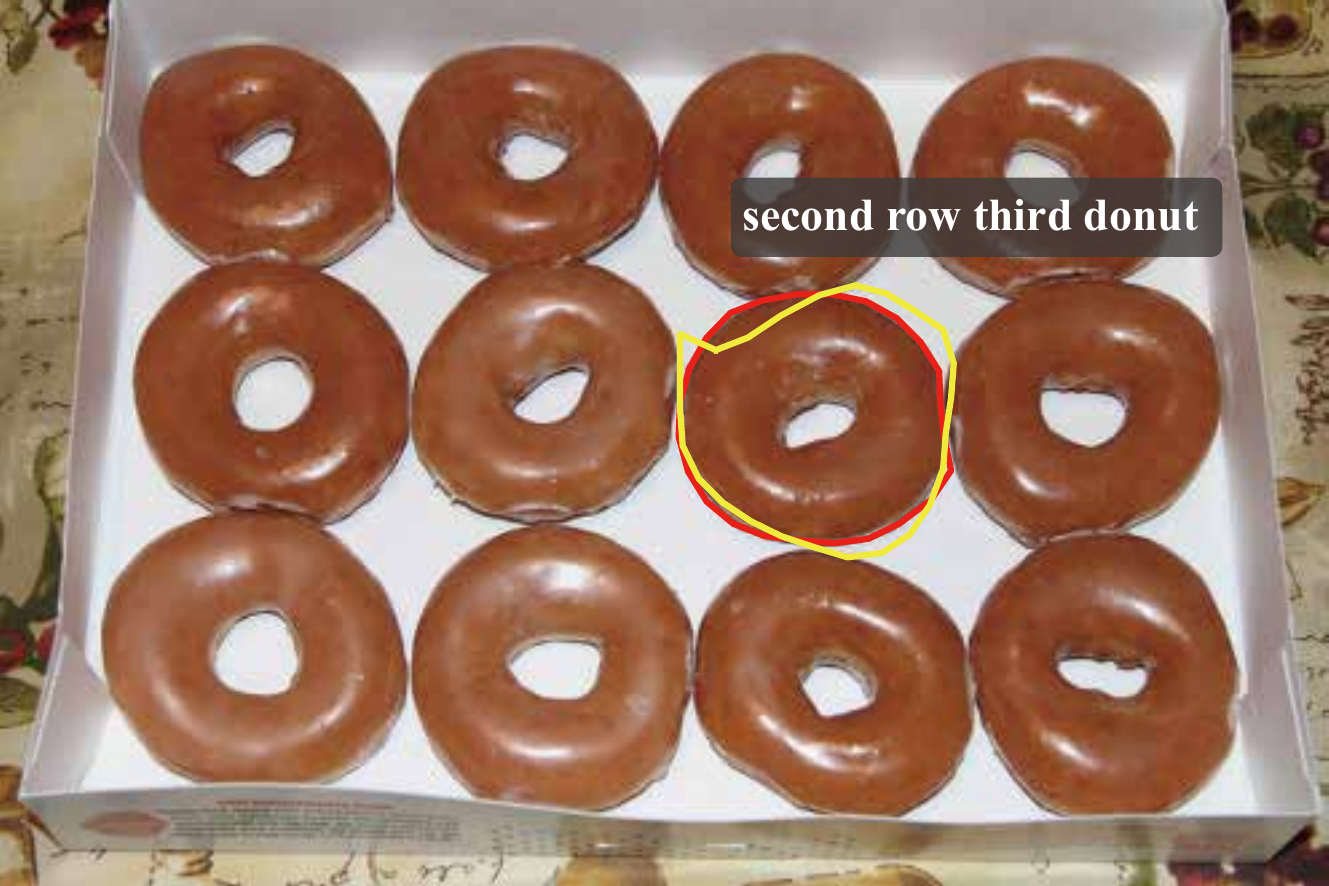}
  \end{subfigure}
  \begin{subfigure}[t]{0.33\textwidth}
  \centering
  \includegraphics[width=0.99\textwidth]{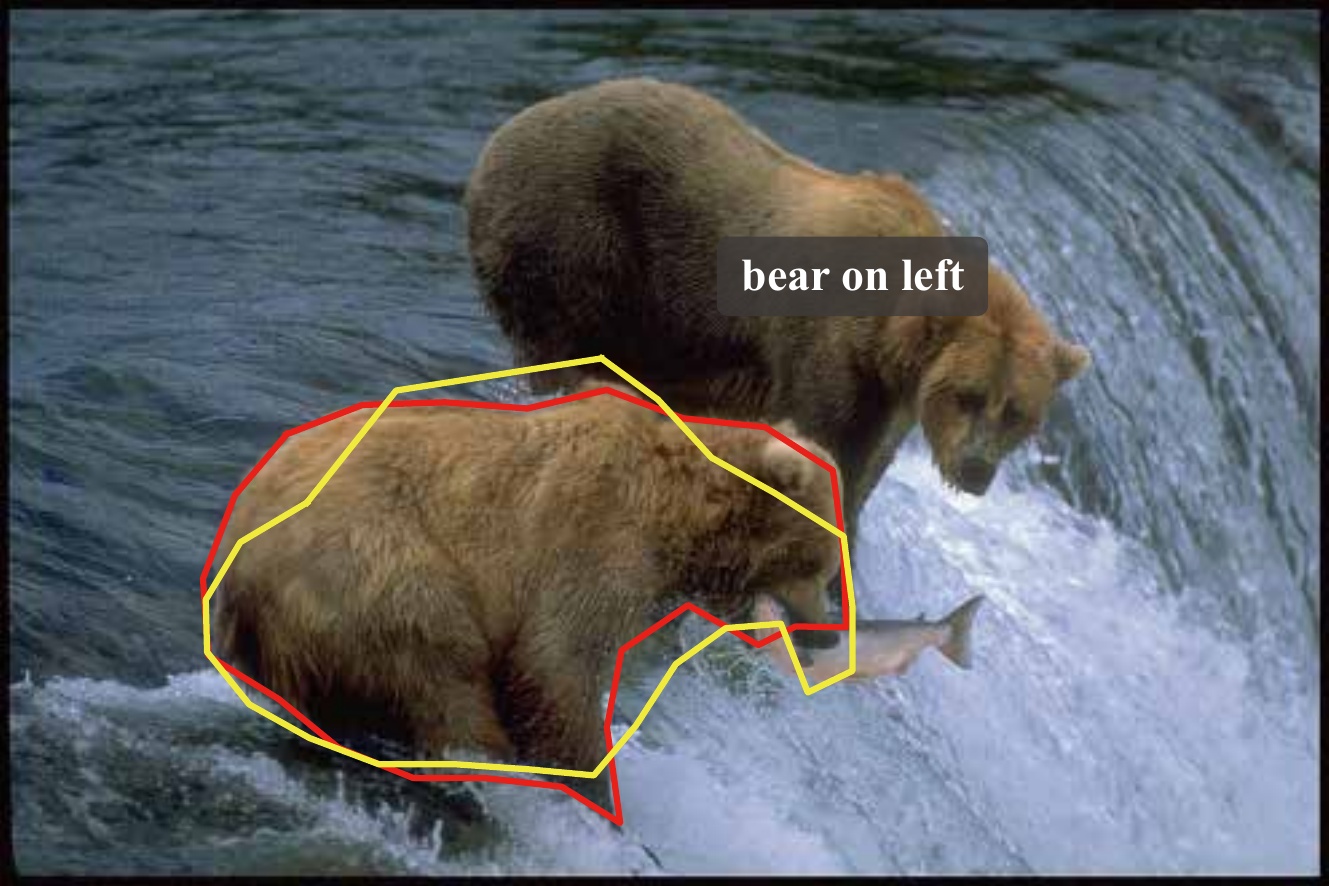}
  \end{subfigure}
  \begin{subfigure}[t]{0.33\textwidth}
  \centering
  \includegraphics[width=0.99\textwidth]{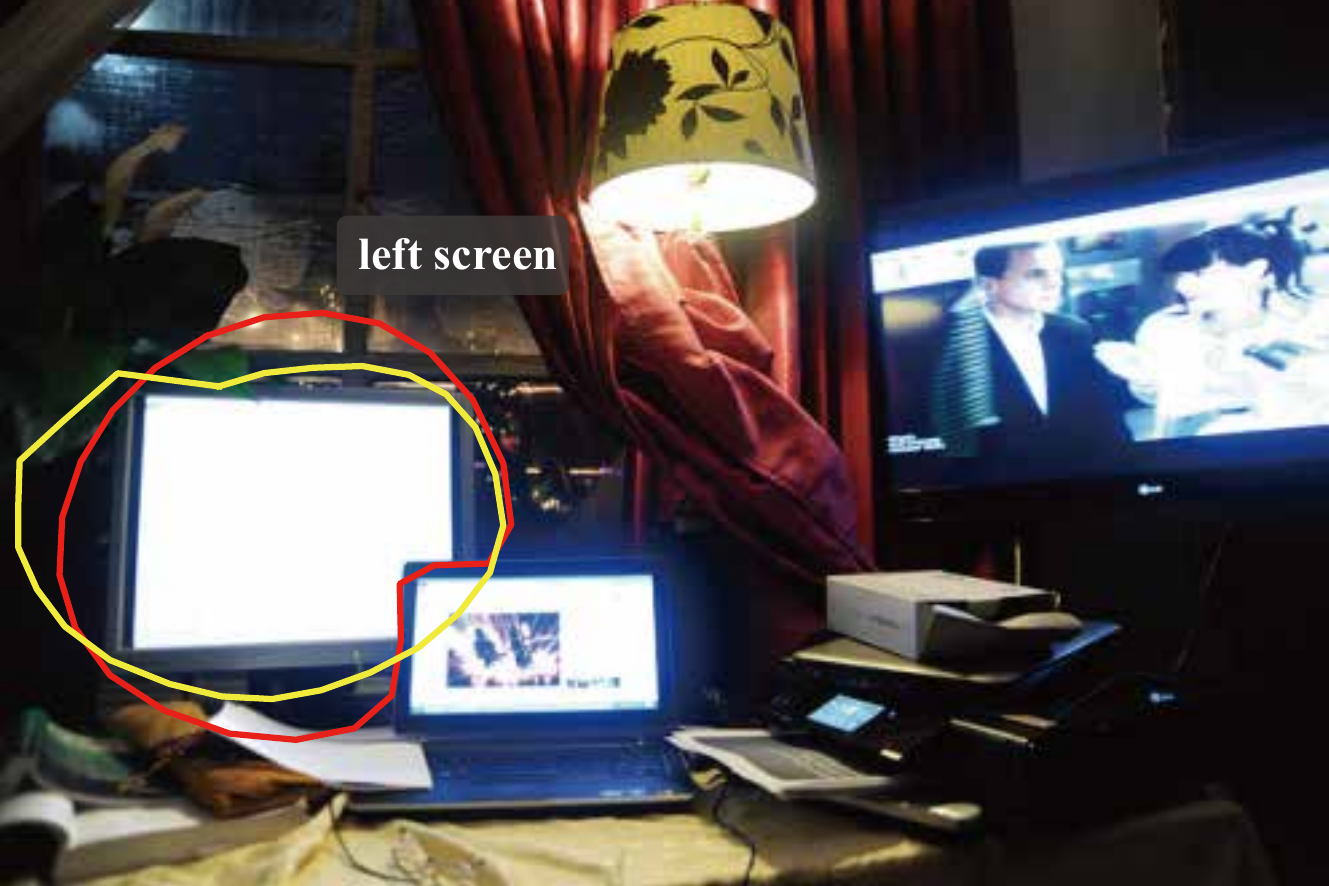}
  \end{subfigure}
  \caption{More Results of referring image segmentation in RefCOCO. The images of the first row are from testA split containing only people, while the second row images from testB consisting of only non-people. The referring expressions of the object are presented in the text box. In the image, the red denotes grounding truth and the yellow symbolizes the prediction.}
  \label{fig:polygon}
\end{figure*}

This section provides more qualitative examples of multiple tasks, including referring bbox detection, and referring image segmentation as illustrated in Fig.\ref{fig:bbox} and Fig.\ref{fig:polygon}, respectively.

\section{Limitation and Future Work}

Our model, leveraging cycle training and multi-task design based on the large language model, exhibited outstanding performance and generalization abilities on a large-scale test set. Nevertheless, there is still room for improvement, particularly in the referring image segmentation task, and the model's generalizability demands further enhancement.

\end{document}